\documentclass{article}

\usepackage{arxiv}
\usepackage{amsmath}

\usepackage[utf8]{inputenc} 
\usepackage[T1]{fontenc}    
\usepackage{hyperref}       
\usepackage{url}            
\usepackage{booktabs}       
\usepackage{amsfonts}       
\usepackage{nicefrac}       
\usepackage{microtype}      
\usepackage{lipsum}
\usepackage{graphicx}
\usepackage{comment}
\graphicspath{ {./images/} }

\DeclareMathOperator*{\argmax}{arg\,max}

\usepackage{appendix}
\usepackage[authoryear]{natbib}

\title{The Cooperative Network Architecture: Learning Structured Networks as Representation of Sensory Patterns}

\author{
  Pascal J. Sager \\
  Zurich University of Applied Sciences\\
  University of Zurich \\
  \texttt{sage@zhaw.ch} \\
   \And
 Jan M. Deriu \\
  Zurich University of Applied Sciences\\
  \texttt{deri@zhaw.ch} \\
  \And
 Benjamin F. Grewe \\
 University of Zurich, ETH Zurich \\
  \texttt{bgrewe@ethz.ch} \\
  \And
 Thilo Stadelmann \\
  Zurich University of Applied Sciences\\
   European Centre for Living Technology (Venice, IT) \\
  \texttt{stdm@zhaw.ch} \\
  \And
 Christoph von der Malsburg \\
  Frankfurt Institute for Advanced Studies \\
  University of Zurich, ETH Zurich \\
  \texttt{malsburg@fias.uni-frankfurt.de} \\
}

\begin{document}
\maketitle
\begin{abstract}
We introduce the \emph{Cooperative Network Architecture (CNA)}, a model that represents sensory signals using structured, recurrently connected networks of neurons, termed ``nets.'' Nets are dynamically assembled from overlapping net fragments, which are learned based on statistical regularities in sensory input. This architecture offers robustness to noise, deformation, and generalization to out-of-distribution data, addressing challenges in current vision systems from a novel perspective. We demonstrate that net fragments can be learned without supervision and flexibly recombined to encode novel patterns, enabling figure completion and resilience to noise. Our findings establish CNA as a promising paradigm for developing neural representations that integrate local feature processing with global structure formation, providing a foundation for future research on invariant object recognition.
\end{abstract}

\keywords{net fragments \and neural code \and pattern recognition \and computer vision \and representation learning}

\section{Introduction}\label{sec:intro}
\begin{figure}[ht!]
  \begin{center}
    \includegraphics[width=\textwidth]{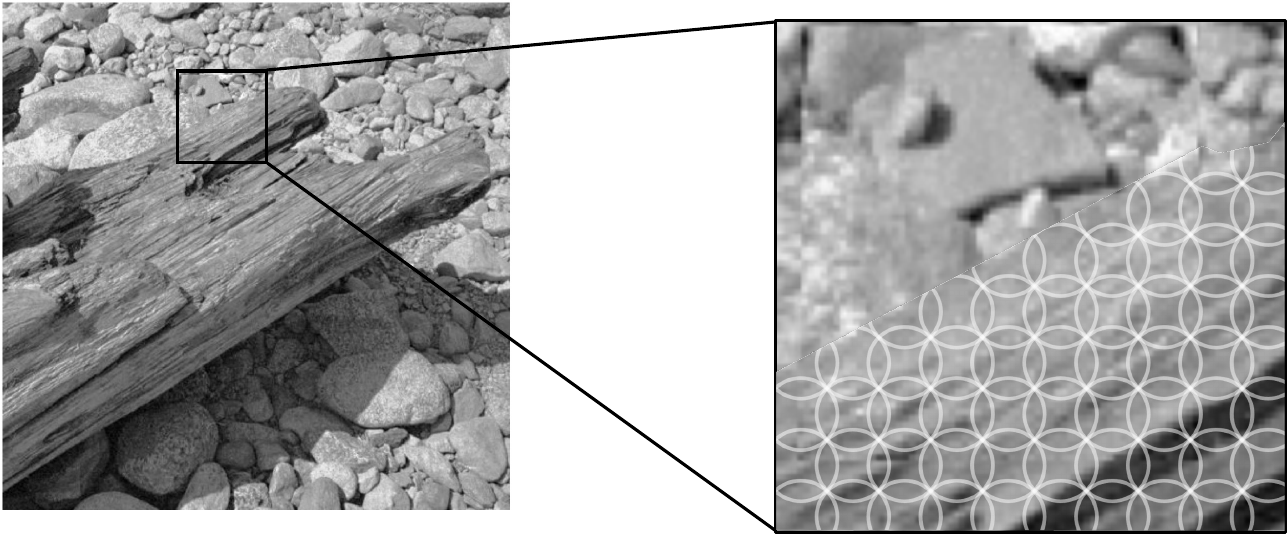}
  \end{center}
  \caption{{\bf Representation of objects by nets.}
Objects are captured in their entirety by coherent nets composed of overlapping net fragments (schematically symbolized by white circles in the right panel).  After training on natural images, each patch of visual space contains a complement of net fragments that represent textures that have been encountered with statistical significance. Net fragments overlap neuron-wise, and those activated by an object coalesce into a coherent net. The conundrum (stated in \cite{Olshausen2005}) that object contours can often not be found with the help of edge detectors due to lack of gray-level contrast (as in places inside the square in the left panel) may be resolved by the idea that contours are defined as borders of the nets covering objects (or covering the background).  Figure adapted from \cite{Olshausen2005}, with permission.}
  \label{fig:bio_net_fragments}
\end{figure}

Artificial intelligence (AI) systems have demonstrated remarkable capabilities in processing diverse types of data, including images, text, and speech \citep{lecun_deep_2015, Sager_2025}.
Yet, despite these advances, artificial vision systems continue to struggle with robust object representation and recognition, particularly under varying viewing conditions, occlusion, or noise \citep{fan2023towards, simmler_survey_2021}.
This limitation highlights a fundamental gap between current AI systems and the flexible, context-aware perception observed in biological vision.

Existing approaches attempt to address these challenges through methods such as feature or sample engineering \citep{amirian2018trace, moosavi-dezfooli_deepfool_2016, meyer2025hounsfield}, domain adaptation \citep{csurka_domain_2017, sager_unsupervised_2022}, and data augmentation \citep{carlucci_hallucinating_2019, tuggener_real_2024}. However, these techniques largely operate within the conventional paradigm of representing objects as static patterns of neuronal activations, often requiring large datasets and extensive supervision to achieve generalization.

In contrast, we propose a new paradigm by introducing the \emph{Cooperative Network Architecture (CNA)}, a model in which object representations emerge from dynamic interactions among recurrently connected neurons. Rather than encoding objects as distinct activation vectors, CNA organizes sensory information into ``nets'' -- structured networks of mutually supporting neurons. These nets are assembled from overlapping ``net fragments,'' which encode frequently co-occurring feature constellations and are acquired through Hebbian plasticity \citep{hebb_organization_1949}. This compositional structure supports robust recognition, denoising, and figure completion through cooperative dynamics without external supervision.

In this paper, we present the first concrete implementation of CNA and demonstrate its application to a simple vision task. Our contributions include
\begin{itemize} \item a formal mathematical definition of nets and net fragments,
\item a computational framework for learning net fragments from sensory input,
\item an illustration of how these fragments can be flexibly composed into object-spanning nets,
\item and an analysis of the emergent filtering and generalization properties of nets.
\end{itemize}

CNA offers the conceptual advance of a biologically inspired computational framework that integrates local and global information processing. Our results indicate that it provides a foundation for more robust and adaptable perception systems, aligning with theoretical proposals for dynamic, network-based neural coding \citep{von_der_malsburg_concerning_2018, von_der_malsburg_theory_2022}.
In this work, we establish the neural code and core computational mechanisms within a single representational area (a layer of neurons in which nets and their constituent fragments are formed, see below).  
Such a single area learns input-driven net fragments that bind together co-occurring sensory features, enabling the assembly of coherent perceptual nets. 
Future architectures will extend this principle to multiple interacting areas, in which additional areas can maintain object-associated net fragments that act as top-down hypotheses. Through reciprocal interactions between input-driven and object-memory areas, such systems would support invariant object recognition and more complex perceptual inference, realizing the broader vision of network-based neural coding models of perception (see our discussion in Section \ref{sec:discussion}).

The remainder of this paper is structured as follows. In Section \ref{sec:fundamentals}, we present the theoretical foundations of our approach and provide the first formal mathematical definition of nets and net fragments, concepts that have previously been described only in prose \citep{von_der_malsburg_concerning_2018, von_der_malsburg_theory_2022}. Section \ref{sec:coop_net_architecture} introduces a concrete computational implementation of this framework. In Section \ref{sec:exp_results}, we report experimental results demonstrating the model’s ability to generalize to out-of-distribution inputs and to perform noise suppression. Section \ref{sec:RelatedWork} situates our approach within the broader literature, contrasting it with related models such as associative memory networks, Boltzmann machines, and modern artificial neural networks. Finally, Section \ref{sec:discussion} discusses the implications of our findings and outlines how the proposed principles could be extended toward multi-area architectures for object recognition.

\section{The Net Fragment Framework}
\label{sec:fundamentals}
Before presenting the computational framework, we introduce its foundational components: neurons, activations, nets, net fragments, and competition. These theoretical definitions serve as the basis for the subsequent sections. Implementation details for the introduced formalisms are provided in the next section.

\paragraph{Neurons.}
A neuron $i$ is denoted by $n_i$. Each neuron $n_i$ has a \emph{local neighborhood}, denoted by the closed ball $\mathcal{B}_r(n_i)$, which is the set of neurons $n_j$ such that the distance $d(n_i, n_j)$ between $n_i$ and $n_j$ is less than or equal to a threshold radius $r>0$, i.e.,
\begin{equation}
\mathcal{B}_r\left(n_i\right) = \left \{n_j \mid d(n_i, n_j) \leq r \right \}.
\label{eq:neighborhood}
\end{equation}
At each time step $t$, the \emph{activity state} of a neuron $n_i$ is represented by $y_{i,t} \in \{0, 1\}$, where $y_{i,t} = 1$ indicates that the neuron is active at $t$. The activity of a neuron over time is characterized by an \emph{activation pattern}, which is a sequence of activation states $\{y_{i,t}\}_{t=1}^T$, where $T$ is the total number of time steps (sequential processing steps).

\paragraph{Connections.}
Neurons interact via \emph{connections}. A direct connection from neuron $n_i$ to another neuron $n_j$ is defined by a positive weight $w_{i,j} > 0$, which facilitates the propagation of neuronal activity from $n_i$ to $n_j$, so that the activity $y_{i,t}$ of neuron $n_i$ can influence the activity $y_{j,t+1}$ of neuron $n_j$ at the subsequent time step. In this work, we limit the connections to local neighborhoods:
\begin{equation}
\forall  w_{i,j} \quad n_j \in \mathcal{B}_r(n_i)
\label{eq:neighborhood_connections}
\end{equation}
Additionally, we limit the weights to a range between $0$ and $1$, allowing neuron $n_i$ only to positively influence $n_j$:%
\begin{equation}
w_{i,j} \in [0,1]
\label{eq:neighborhood_connections_range}
\end{equation}
We define a relation $n_i \sim n_j$ between two neurons to indicate that their activation patterns are highly correlated, i.e., $y_{i,t}$ often influences  $y_{j,t+1}$.
Under Hebbian learning \citep{hebb_organization_1949}, the connection strength $w_{i,j}$ increases when there is $n_i \sim n_j$.

\paragraph{Net Fragments.}
A \emph{net fragment}  is defined as the set of neurons that provide direct activation support to a given neuron.
A neuron $n_j$ is said to provide \emph{support} to another neuron $n_i$ if two conditions are met:
(1) there exists a high correlation between two neurons $n_j$ and $n_i$ such that $n_j \sim n_i \Rightarrow w_{ji} > 0$; and (2) neuron $n_j$ was active at the previous time step, i.e., $y_{j,t-1}=1$.
Since synaptic weights are bounded within the interval $w_{j,i} \in [0,1]$, the degree of support a neuron can provide is likewise restricted to this range.

Formally, the \emph{net fragment} associated with neuron $n_i$ at time $t$ is given by
\begin{equation}
  \mathcal{F}_{t}(n_i) = \{n_i\} \cup \{n_j \mid w_{j,i} > 0 \land y_{j,t-1} = 1\}.
  \label{eq:net_fragments}
\end{equation}
In this formulation, any neuron $n_j \in \mathcal{F}(n_i)$  contributes to the activation likelihood of $n_i$ at time $t$.
Notably, neuron $n_i$ is part of its own fragment, reflecting a form of self-support -- if active at time $t$, the neuron may contribute to sustaining its own activation at the subsequent time step.
Note that $n_i \sim n_j$ is not an equivalence relation since correlation is not transitive, meaning that one neuron can be part of multiple net fragments.

\paragraph{Activation Function.}
The decision of whether a neuron becomes active is based on the amount of support it receives from within the net fragment. We call this accumulated input the \emph{pre-synaptic activity} at timestep $t$, which is a function $\mathcal{I}$ of the net fragment $\mathcal{F}_{t-1}(n_i)$ of a neuron $n_i$ at timestep $t-1$. 
\begin{equation}
  \boldsymbol{a}^\mathcal{I}_{i, t} = \mathcal{I}\left(\mathcal{F}_{t}(n_i) \right)  
  \label{eq:pre_synaptic_activity}
\end{equation}
The pre-synaptic activity $a^\mathcal{I}_{i, t} \in \mathbb{R}$ is large when many other neurons provide support.
Since supporting neurons exhibit connections that have been selectively strengthened through previous mutual activity and Hebbian updates within specific contexts, their activity effectively serves to confirm the contextual plausibility of a given neuron's activation.
The very general formulation of $\mathcal{I}$ shows that the pre-synaptic activity can be calculated with any desired function.
In this work, we specifically define $\mathcal{I}$ as the weighted sum of prior activations from supporting neurons:
\begin{equation}
\boldsymbol{a}^\mathcal{I}_{i, t} = \sum_{n_j \in \mathcal{F}_{t}(n_i)} \boldsymbol{w}_{j,i} \cdot \boldsymbol{y}_{j,t-1}.
\label{eq:specific_pre_synaptic_activity}
\end{equation}

The activation of a neuron at timestep $t$ is determined by the activation function $\mathcal{A}$ that depends on the pre-synaptic activity $a_{i, t}$ and environmental conditions $\mathcal{E}$: 
%
\begin{equation}
  \boldsymbol{y}_{i, t} = \mathcal{A} \left(\boldsymbol{a}_{i, t}, \mathcal{E} \right) \in \{0,1\}
  \label{eq:activation_framework}
\end{equation}
The environmental conditions $\mathcal{E}$ include an \emph{attenuation} signal, a constraint that selectively suppresses weak excitatory inputs by requiring neurons to accumulate a sufficiently large pre-synaptic activity before firing.
Again, we here provide a very general formulation, using $\mathcal{A}$ and $\mathcal{E}$, to emphasize that any activation function or environmental mechanism can be used.
In this work, we instantiate $\mathcal{E}$ using a combination of normalization (see Appendix~\ref{sec:normalization}) and attenuation (see Eq.~\ref{eq:inhibition}), and define $\mathcal{A}$ as the heaviside step function (see Eq.~\ref{eq:s2_final_output}). 

\paragraph{Nets.}
Neurons are not only influenced by direct connections within net fragments but also by indirect connections (e.g., if $n_i$ supports $n_j$ and $n_j$ supports $n_k$, then $n_i$ indirectly supports $n_k$).
A set of directly and indirectly supported neurons is called a \emph{net}, implementing the basic structure used in this work to represent objects.
For simplicity in the following definitions, we introduce $\mathcal{F}^*_{t}(n_i)$, a net fragment $\mathcal{F}_{t}(n_i)$ in which $n_i$ is active as well:
\begin{equation}
  \mathcal{F}^*_{t}(n_i) =\mathcal{F}_{t}(n_i) \quad \text{with} \quad  y_{i, t}=1
  \label{eq:net_fragments_full}
\end{equation}
To account for both direct and indirect influences on a neuron's activation, we use a recursive formulation of nets from the perspective of a single neuron. In the base case, a net $\mathcal{N}_t^{1}[n_i]$ is defined as the immediate net fragment:
\begin{equation}
    \mathcal{N}_t^{1}[n_i]  = \mathcal{F}^*_t(n_i)
\end{equation}
For $r \geq 2$, the recursive relation is given by
\begin{equation}
    \mathcal{N}_t^{r}[n_i]  = \mathcal{N}_t^{r-1}[n_i] \bigcup_{n_j \in \mathcal{N}_t^{r-1}[n_i]} \mathcal{F}^*_t(n_j)
    \label{eq:recursive_net_fragments}
\end{equation}
The complete network accessible from $n_i$ is then defined as the closure
\begin{equation}
    \mathcal{N}_t^{*}[n_i]  = \bigcup_{r=1}^{\infty} \mathcal{N}_t^{r}[n_i],
\end{equation}
which comprises all neurons in the active subgraph that are reachable from $n_i$.
This recursive formulation illustrates how support relationships propagate through time: local net fragments $\mathcal{F}_t(n_i)$, which reflect immediate sources of support for a neuron, are combined recursively to form increasingly larger subnetworks $\mathcal{N}_t^{r}[n_i]$. These progressively include neurons with indirect but recurrent influence on $n_i$'s activation. This process reflects how localized evidence (elementary activations) can aggregate into a coherent larger pattern (composite activation), corresponding to a higher-order feature representation in the network.
Moreover, the explicit time dependence in these definitions underscores the dynamic nature of both the net fragments and the overall net, as they are continuously reassembled in response to the evolving support received from neighboring neurons, and increasing attenuation that prevents neurons with a low pre-synaptic activity from firing (see next section).

In the proposed architecture, no symmetry constraint is imposed on the weights, that is, $w_{i,j} \neq w_{j,i}$ in general.
However, due to the temporal dynamics of the network, activations tend to stabilize over a longer time horizon $T$, such that for sufficiently large $t$, the activations become approximately stationary, i.e., $y_{i, t} \approx y_{i, t+1}$.
Under these conditions, the mutual influence between neurons becomes symmetric, meaning that $n_i \sim n_j$ implies $n_j \sim n_i$, and almost symmetric weights emerge naturally over time, even in the absence of explicit symmetry constraints.

As a result, it can be assumed that for any pair of distinct neurons $n_i \neq n_j$, their closures either converge to the same set of neurons or remain largely disjoint.
Formally, this can be expressed as:
\begin{equation}
    \mathcal{N}_t^{*}[n_i]  \approx \mathcal{N}_t^{*}[n_j] \quad \text{or} \quad  \mathcal{N}_t^{*}[n_i]  \cap  \mathcal{N}_t^{*}[n_j] \approx \emptyset.
\end{equation}

\paragraph{Competitive Neurons.}
One issue with the current formulation is that the mapping
\begin{equation}
  \mathcal{A}\circ\mathcal{I}:2^{\mathcal{N}}\;\longrightarrow\;\{0,1\}
\end{equation}
treats any two net fragments that yield the same activation identically.  In particular, we say
\begin{equation}
  \mathcal{F}_{t_k}(n_i)\simeq\mathcal{F}_{t_l}(n_i)
  \quad\Longleftrightarrow\quad
  \mathcal{A}\bigl(\mathcal{I}(\mathcal{F}_{t_k}(n_i)),\mathcal{E}\bigr)
  =\mathcal{A}\bigl(\mathcal{I}(\mathcal{F}_{t_l}(n_i)),\mathcal{E}\bigr),
\end{equation}
meaning different contexts collapse to the same output for $n_i$. To break this symmetry and make each neuron context-aware, we introduce \emph{competitive neurons}.

For each neuron $n_i$, We introduce a set of alternatives $\{n_i^1,\dots,n_i^\kappa\}$, all located at the same spatial position $\left ( \mathcal{B}_r(n_i)=\mathcal{B}_r(n_i^k) \right )$.
These alternatives undergo competition and one of them is permitted to be active within a given context. This exclusivity encourages these competitive neurons to specialize, over time, into representations of different contextual variations.

To formalize these competitive neurons, we first define the \emph{competition class} of a neuron $n_i$:
\begin{equation}
  \mathcal{C}(n_i) 
  = \bigl\{ n_i^k  \, \text{ where } \, \mathcal{B}_r(n_i)=\mathcal{B}_r(n_i^k) \bigr\},
\end{equation}
These neurons are mutually exclusive candidates to represent the context-dependent expression of a feature at a fixed spatial position.
During each time step, after computing the pre-synaptic activities $\boldsymbol{a}_{j,t}^\mathcal{I}$ for $n_j\in\mathcal{C}(n_i)$, we select a small winner set
\begin{equation}
  W_t(n_i)
  = \mathrm{Comp}\bigl(\{\,\boldsymbol{a}_{j,t}^\mathcal{I}\mid n_j\in\mathcal{C}(n_i)\}\bigr).
\end{equation}
While various rules to determine the winning set can be used, we apply winner-takes-all selection in this work (for implementation details, see Section~\ref{sec:stage_2_impl}).
Only neurons in $W_t(n_i)$ are eligible to fire:
\begin{equation}
  y_{i,t} \;=\;
  \begin{cases}
    \mathcal{A}\bigl(\boldsymbol{a}_{i,t}^\mathcal{I},\mathcal{E}\bigr)
      & \text{if }n_i\in W_t(n_i),\\
    0 & \text{otherwise.}
  \end{cases}
\end{equation}

By enforcing mutual exclusivity among neurons with overlapping receptive fields, competition ensures that different contexts activate distinct neuronal subsets.

\section{Cooperative Network Architecture}\label{sec:coop_net_architecture}

\begin{figure}[ht!]
  \begin{center}
    \includegraphics[width=\textwidth]{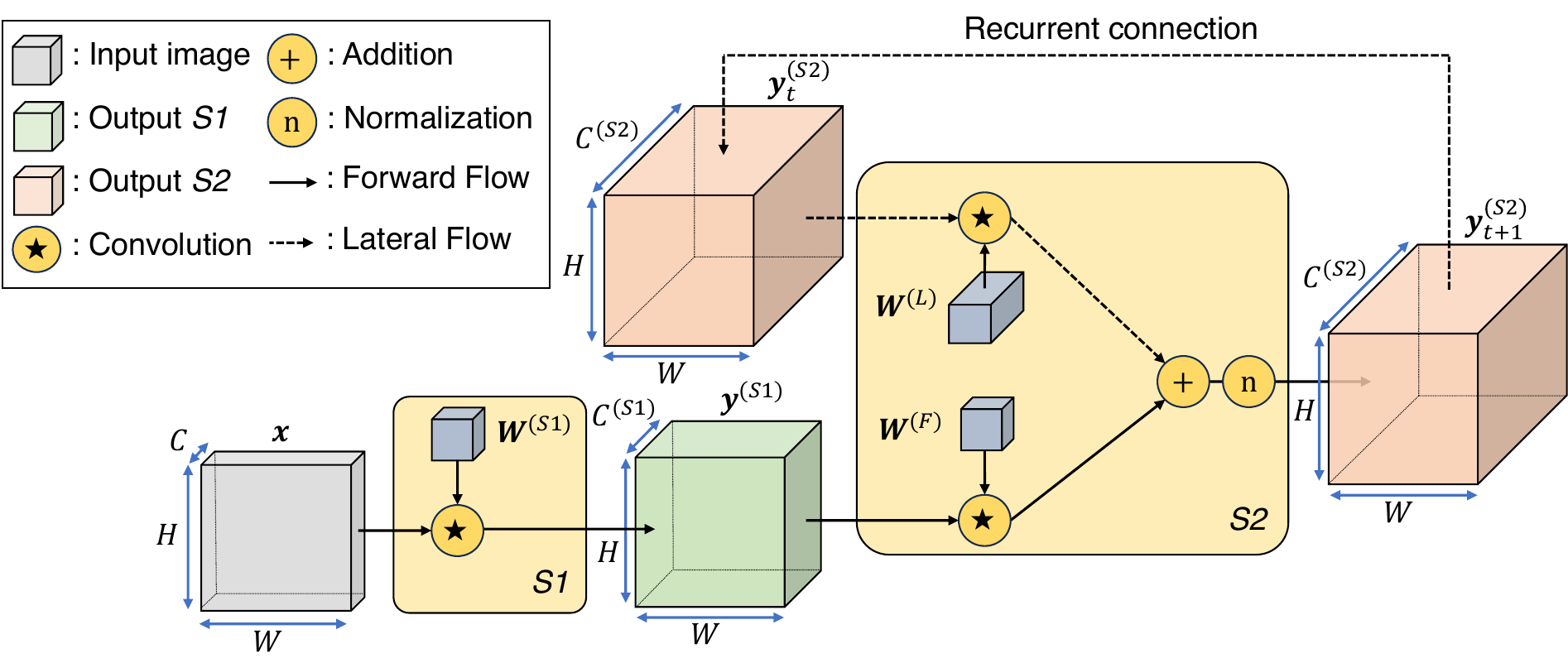}
  \end{center}
  \caption{\textbf{Dynamics within the proposed CNA}: The input image is fed into stage $S1$ to obtain feature activations. The feature activations are then processed by stage $S2$, together with activations from neurons within the same layer. 
  }\label{fig:system_dynamics}
\end{figure}

In this section, we present a concrete instantiation of the proposed framework, specifically applied to the visual domain -- although the framework is general and may be extended to other sensory modalities in the future. We refer to this implementation as the Cooperative Network Architecture (CNA).
To realize our framework, we adopt a standard convolutional neural network (CNN) architecture~\citep{lecun_backpropagation_1989}.

The CNA comprises two processing stages, depicted as yellow squares in Figure~\ref{fig:system_dynamics}. \emph{Stage 1 ($S1$)} performs fixed (i.e., not learned) feature extraction directly from the image input. This fixed feature stage is used here for simplicity and to facilitate clearer diagnostics of the subsequent results by combining well-characterized, predefined features. However, these features could also be learned in future work, for example, through biologically inspired Hebbian learning schemes that have been shown to induce convolutional, Gabor-like filters \citep[e.g.,][]{miconi_hebbian_2021}. \emph{Stage 2 ($S2$)} implements the core principles of the described computational framework by learning net fragments.

The network dynamics unfold over discrete time steps $t = [0, \dots, T]$.
Since the feature extraction stage ($S1$) employs four fixed filters, it yields a time-invariant activation pattern for a given (fixed) input.
In contrast, the second stage ($S2$) exhibits dynamic temporal evolution, where internal representations are iteratively refined at each time step ($t \to t+1$).
Specifically, we use a progressively increasing global attenuation signal that only allows neurons with a high pre-synaptic activity to remain active.
The following sections discuss these two processing stages.

\subsection{Stage 1: Feature Extraction}
Stage 1 is designed to model the sensory input by extracting initial neural activations from the raw image data. This extraction is essential due to the continuous nature of typical image inputs (e.g., grayscale or RGB), whereas the computational framework employed in this work operates on binary activations. Consequently, this stage serves as a preprocessing step, enabling the transformation of continuous-valued images into a suitable binary representation.

The input image is denoted as $\boldsymbol{x} \in \mathbb{R}^{C, H, W}$ (illustrated by the gray box in Figure \ref{fig:system_dynamics}), where $C$ indicates the number of image channels ($1$ for grayscale, $3$ for RGB), while $H,  W$ stand for the height and width of the image.

\paragraph{Activation Computation.}
To extract initial activation patterns, we use four fixed, non-trainable filters (details see Appendix \ref{sec:features}). These filters are implemented as standard convolutional kernels, used to obtain the binary activation pattern $\boldsymbol{y}^{(S1)} \in \{0,1\}^{C^{(S1)}, H, W}$ as follows:

\begin{equation}
\boldsymbol{y}^{(S1)}_{c_{\text{out}}} = \Theta \left(\sum_{c_{\text{in}}=0}^C \boldsymbol{W}^{(S1)}_{c_{\text{out}},c_{\text{in}}} \star \boldsymbol{x}_{c_{\text{in}}} \right)
\label{eq:y_s1}
\end{equation}

Here, $\star$ denotes the convolution operator between the filters $\boldsymbol{W}^{(S1)} \in \mathbb{R}^{C^{(S1)}, C, h, w}$ and the input image $\boldsymbol{x}$, while the parameters  $h, w$ define the width and height of the convolutional filters, i.e., the spatial neighborhood from where the features are extracted.
To binarize the activations, we use the heaviside function $\Theta$, which sets a neuron to one if its input is larger than zero, and to zero otherwise.
\begin{equation}
  y_{i} = \Theta(\boldsymbol{a}_{i} ) = \begin{cases}
      1 \quad \text{if } \boldsymbol{a}_{i}  > 0\\
      0 \quad \text{otherwise}
  \end{cases}
  \label{eq:activation}
\end{equation}
Note that the width and height between input and output activations remain identical, but the number of channels changes from $C$ to $C^{(S1)}$, allowing the extraction of multiple features at each spatial location.

\subsection{Stage 2: Net Fragments}\label{sec:stage_2_impl}
$S2$ constructs net fragments from the binary feature activations $\boldsymbol{y}^{(S1)}$ of $S1$ and integrates them into coherent nets over time. To ensure stability, single neurons or small groups of neurons are suppressed through an attenuation mechanism, ensuring that only large-scale, consistent activation patterns persist. 

\paragraph{Network Topology.}
Stage 2 uses convolutional layers to implement the connections $w_{j,i}$ within net fragments $n_j \in \mathcal{F}(n_i)$.
Conceptually, this stage comprises a three-dimensional array of neurons with dimensions $C^{(S2)}\times H \times W$, where $C^{(S2)}$ denotes the number of feature channels in $S2$, and $H$ and $W$ correspond to the spatial dimensions of the input image.

The spatial distance between two neurons $n_i$ and $n_j$ is defined using the maximum norm (also known as Chebyshev distance), where adjacent neurons have a unit distance:

\begin{equation}
d(n_i,n_j) = \max \left( |h_i - h_j|, |w_i - w_j|\right)
\end{equation}

Each neuron $n^{(S2)}_i$ in stage 2 is connected in two ways: (1) \emph{Feedforward connections} to the corresponding local neighborhood $\mathcal{B}_r(n^{(S1)}_i)$ in $S1$, and (2) \emph{lateral connections} to neighboring neurons within stage 2 itself, $\mathcal{B}_r(n^{(S2)}_i)$, denoted as:

\begin{equation}
\begin{aligned}
    \mathcal{B}_r \left(n^{(S2)}_i \right) &= \mathcal{B}_r \left(n^{(S1)}_i \right) \cup \left\{n_j^{(S2)} \mid d(n_i^{(S2)}, n_j^{(S2)}) \leq r \right\}\\
    \mathcal{B}_r \left(n^{(S1)}_i \right) &= \left\{n_j^{(S1)} \mid d(n_i^{(S2)}, n_j^{(S1)}) \leq r \right\}
\end{aligned}
\end{equation}

where $r$ defines the radius of lateral connections. The convolutional architecture realizes these neighborhood structures via convolutional filters, where the filter size directly determines the radius of both feedforward and lateral connections.
Using convolutional filters not only implements the connectivity restrictions to local neighborhoods but also allows detection of the same pattern at various positions, enabling learning of net fragments independent of position.

The forward connections from $S1$ to $S2$, denoted as $\boldsymbol{W}^{(F)}$, are necessary to activate cells within $S2$. It can be interpreted as a small group of neurons in $S1$ (based on the filter size of $\boldsymbol{W}^{(F)}$) supporting a single neuron within $S2$.
In contrast, the lateral recurrent connections within $S2$, denoted as $\boldsymbol{W}^{(L)}$, are responsible for building the overlapping net fragments within $S2$, providing support amongst the neurons within $S2$.
It is important to note that the lateral connection matrix $\boldsymbol{W}^{(L)}$ does not implement recurrent processing in the traditional sense of sequential token-wise updates, as found in  RNN-based language models. Instead, $\boldsymbol{W}^{(L)}$ iterates over the same neuronal state vector $\boldsymbol{y}^{(S2)}$.
Specifically, the lateral connections $\boldsymbol{W}^{(L)}$ calculate the state updates of $\boldsymbol{y}^{(S2)}_{t\rightarrow t+1}$ within an iterative refinement process in which individual neuronal activations ${y}_i^{(S2)} \in \boldsymbol{y}^{(S2)}$ may flip due to changes in support and attenuation.

\paragraph{Pre-Synaptic Activity.}
To determine the support a single neuron $n_{i,t}$ receives from the net fragment $\mathcal{F}_{t-1}(n_{i, t-1})$, we first calculate the pre-synaptic activity $\mathcal{I}$ by applying convolutional operations on the forward and lateral connections and then taking the sum of these operations: 
\begin{equation}
\begin{aligned}
\boldsymbol{a}^{\mathcal{I}(S2)}_{c_{\text{out}}, t}
&= \underbrace{
    \sum_{c_\text{in}=0}^{C^{(S1)}} \boldsymbol{W}^{(F)}_{c_{\text{out}}, c_{\text{in}}} 
    \star \boldsymbol{y}_{c_{\text{in}}}^{(S1)}
}_{\textit{forward input } (\boldsymbol{a}^{\mathcal{I}(F)})}
+ 
\underbrace{
    \sum_{c_\text{in}=0}^{C^{(S2)}} \boldsymbol{W}^{(L)}_{c_{\text{out}}, c_{\text{in}}} 
    \star \boldsymbol{y}_{c_{\text{in}}, t-1}^{(S2)}
}_{\textit{lateral input } (\boldsymbol{a}^{\mathcal{I}(L)})}
\\[1.5ex]
\boldsymbol{a}^{\mathcal{I}(F)}_{c_{\text{out}}}
&= \sum_{c_\text{in}=0}^{C^{(S1)}} \sum_{h=0}^{H-1} \sum_{w=0}^{W-1} 
  \boldsymbol{W}^{(F)}_{c_{\text{out}}, c_{\text{in}}, h, w} \cdot 
  \boldsymbol{y}^{(S1)}_{c_{\text{in}}, h'+h, w'+w}
\\
\boldsymbol{a}^{\mathcal{I}(L)}_{c_{\text{out}}, t}
&= \sum_{c_\text{in}=0}^{C^{(S2)}} \sum_{h=0}^{H-1} \sum_{w=0}^{W-1} 
  \boldsymbol{W}^{(L)}_{c_{\text{out}}, c_{\text{in}}, h, w} \cdot 
  \boldsymbol{y}^{(S2)}_{c_{\text{in}}, h'+h, w'+w, t-1}
\end{aligned}
\label{eq:a_a2}
\end{equation}
Here, the forward connections have the shape $\boldsymbol{W}^{(F)} \in \mathbb{R}^{C^{(S2)}, C^{(S1)}, h, w}$ and the recurrent connections have the shape $\boldsymbol{W}^{(L)} \in \mathbb{R}^{C^{(S2)}, C^{(S2)}, h, w}$, where $C^{(S2)}$ is the number of channels in $S2$. 
Therefore, the pre-synaptic activity $\mathcal{I}$ at time $t$ is increased by the learned amount $\boldsymbol{w}_{j,i}$ from all connected and active neurons ($\boldsymbol{y}_{j,t-1} = 1$).
Higher support, indicated by larger $\boldsymbol{a}^{\mathcal{I}(S2)}_{i}$, suggests a more likely pattern, as neurons encoding frequent patterns gain reinforcement not only from feedforward connections but also from lateral connections.

Next, we extend this convolutional-based formulation of forward and lateral connections by introducing competitive neurons.
Afterward, we describe how the pre-synaptic activity $\boldsymbol{a}^{\mathcal{I}(S2)}$ undergoes attenuation, how it is transformed into $\boldsymbol{y}^{(S2)}$,
and how the weight matrices $\boldsymbol{W}^{(F)}, \boldsymbol{W}^{(L)}$ are initialized and trained.

\paragraph{Competitive Neurons.}
To address the challenge of context entanglement, wherein individual neurons may participate in multiple fragments, we implement \emph{competitive neurons} as follows: For each base feature channel $c \in C^{(S2)}$, we introduce $\kappa$ parallel alternatives that share the same receptive field structure but are allowed to develop distinct connections. Following our formal definition, each base neuron $n_{c,h',w'}$ at spatial location $(h',w')$ has the competitive class:

\begin{equation}
    \mathcal{C} \left( n_{c,h',w'} \right) = \left\{ n_{c,h',w'}^1, \dots, n_{c,h',w'}^\kappa\right\}
\end{equation}

We implement this by introducing an additional competition dimension of size $\kappa$ to the tensors in $S2$:
\begin{equation}
\begin{aligned}
   C^{(S2)} &\to (C^{(S2)},\,\kappa)\\
   \boldsymbol{W}^{(F)}\in\mathbb R^{C^{(S2)},\,C^{(S1)},\,h,\,w}
   &\to
   \mathbb R^{C^{(S2)},\,\kappa,\,C^{(S1)},\,h,\,w}\\
   \boldsymbol{W}^{(L)}\in\mathbb R^{C^{(S2)},\,C^{(S2)},\,h,\,w}
   &\to
   \mathbb R^{C^{(S2)},\,\kappa,\,C^{(S2)},\,h,\,w}\\
   \boldsymbol{a}^{\mathcal{I}(L)}\in\mathbb R^{C^{(S2)},H,W}
   &\to
   \mathbb R^{C^{(S2)},\,\kappa,\,H,W}
\end{aligned}
\end{equation}

Although this transformation increases the number of total feature maps by a factor of $\kappa$, only a single competitive neuron per base channel is active at any given spatial location (one neuron within $\mathcal{C} \left( n_{c,h',w'} \right)$). This sparsity is enforced through a group-wise \emph{winner-take-all (WTA)} mechanism along the $\kappa$-dimension.

At each time step $t$, competition operates independently within each class $\mathcal{C} \left( n_{c,h',w'} \right)$. For each base neuron, we compute pre-activation scores for all competitive neurons and select the winning competitive index:
\begin{equation}
  k^*(c,h',w') \;=\;
  \argmax_{k \in \{0,\dots,\kappa-1\}}\;
    \left\langle
      \boldsymbol{W}_{c,k,:,:,:,:},\;
      \boldsymbol{z}_{h',w',t}
    \right\rangle.
\end{equation}

where $\boldsymbol{z}_{h',w',t}$ is the concatenated input patch:

\begin{equation}
\boldsymbol{z}_{h',w',t} \;=\;
\begin{bmatrix}
\boldsymbol{y}^{(S1)}_{:,\, h':h'+H,\, w':w'+W} \\
\boldsymbol{y}^{(S2)}_{:,\, h':h'+H,\, w':w'+W,\, t-1}
\end{bmatrix}.
\end{equation}
Thus, at each spatial location and time step, the index $k^* \in \{1, \dots, \kappa \}$ maximizing the correlation between the corresponding filter and the local input patch is chosen.
Once the winning index $k^*$ is selected, the activation is computed exclusively for the corresponding neuron:
\begin{equation}
\begin{aligned}
\boldsymbol{a}^{\mathcal{I}(S2)}_{c, h', w', t} &=
\boldsymbol{a}^{\mathcal{I}(F)}_{c, k^*, h', w'} + \boldsymbol{a}^{\mathcal{I}(L)}_{c, k^*, h', w', t}
\\[1.5ex]
\boldsymbol{a}^{\mathcal{I}(F)}_{c, k^*, h', w'} &= 
\sum_{c_{\text{in}}=0}^{C^{(S1)} - 1} \sum_{h=0}^{H-1} \sum_{w=0}^{W-1}
\boldsymbol{W}^{(F)}_{c,k^*,\, c_{\text{in}},\, h,\, w} \cdot
\boldsymbol{y}^{(S1)}_{c_{\text{in}},\, h'+h,\, w'+w}
\\[1.5ex]
\boldsymbol{a}^{\mathcal{I}(L)}_{c, k^*, h', w', t} &= 
\sum_{c_{\text{in}}=0}^{C^{(S2)} - 1} \sum_{h=0}^{H-1} \sum_{w=0}^{W-1}
\boldsymbol{W}^{(L)}_{c,k^*,\, c_{\text{in}},\, h,\, w} \cdot
\boldsymbol{y}^{(S2)}_{c_{\text{in}},\, h'+h,\, w'+w,\, t-1}
\end{aligned}
\end{equation}

Initially, all competitive neurons $n_{c,h',w'}^k$ are identical due to shared initialization. However, Hebbian learning induces specialization: each competitive neuron within $\mathcal{C} \left( n_{c,h',w'} \right)$ gradually diverges, learning to encode different contextual variations of the same base feature.

In a practical implementation, we merge the $\kappa$ and channel dimensions, resulting in a flat channel axis of size $C^{(S2)} \cdot \kappa$. This allows the use of efficient standard 3D convolutional layers, where non-winning neurons are simply masked out at each spatial location by setting them to $0$, preserving the theoretical definition.

\paragraph{Attenuation.}
The pre-synaptic activity is processed by an activation function $\mathcal{A} \left(\boldsymbol{a}^\mathcal{I}, \mathcal{E} \right)$. In our implementation, this function first normalizes the potential  $\boldsymbol{a}^\mathcal{I} \rightarrow \boldsymbol{a}^{\text{norm}}$ to conform it into the range $[0, 1]$ (details see Appendix \ref{sec:normalization}), and then applies a globally increasing attenuation pressure $\boldsymbol{a}^{\text{norm}} \rightarrow \boldsymbol{a}$ to ensure that only neurons receiving substantial support remain active.
Specifically, attenuation is implemented as:
\begin{equation} 
    \boldsymbol{a}^{(S2)}_{c,t} = \left(\boldsymbol{a}^{\text{norm }(S2)}_{c,t}\right)^\gamma,
    \label{eq:inhibition}
\end{equation}
where $\gamma$ regulates the attenuation strength and increases over time according to $\gamma=\alpha+\beta t$ (when not stated differently, we set $\alpha=1.2$ and $\beta=0.2$), making activation progressively harder. This process leads to a sparsification of activations that occurs through both direct attenuation and reduced support from previously silenced neurons~.

The final output is then binarized using the heaviside function $\Theta$:
\begin{equation} 
     \boldsymbol{y}_{c,t}^{(S2)} = \Theta\left(\boldsymbol{a}^{(S2)}_{c,t}  - b^{(S2)} \right)
     \label{eq:s2_final_output}
\end{equation}
The activation at the last timestep $\boldsymbol{y}^{(S2)}_{t=T}$ represents a recurrently refined activation pattern where unsupported neurons have been suppressed, yielding a coherent representation.

\paragraph{Weight Initialization.}
The forward connections $\boldsymbol{W}^{(F)}$ are initialized so that the activations of the $S1$ neurons are initially copied into all competitive $S2$ neurons in the same position. This is done by setting the connections at the center of the kernel (at position $(h/2, w/2)$) to $1$ if $c_{in} = \lceil c_{out} / \kappa \rceil$, where $c_{in}$ denotes the index of the input channel, $c_{out}$ the index of the output channel, and $\kappa$ the number of competitive neurons. The recurrent connections $\boldsymbol{W}^{(L)}$ are initialized with zeros except for the self-coupling of neurons (recurrent connection to itself), which is set to $1$.

\paragraph{ Learning of Weights.}
Support within net fragments is \emph{learned} via Hebbian plasticity, so that frequently occurring patterns that lead to correlated activation patterns between neurons $n_i \sim n_j$ are ``stored'' in the system and receive more support:
\begin{equation}
    \boldsymbol{W}^{(S2)} := \min \left( \max \left( \boldsymbol{W}^{(S2)} + \eta \cdot \boldsymbol{\rho}_{avg}, 0\right), 1\right)
    \label{eq:hebbian_plasticity}
\end{equation}
Here, $\eta$ represents the learning rate and $\boldsymbol{W}^{(S2)}$ stands for both $\boldsymbol{W}^{(F)}$ and $\boldsymbol{W}^{(L)}$.
Note that the same convolutional kernel is applied at various spatial positions. The decision to update this connection is based on the average correlation $\boldsymbol{\rho}_{avg}$ between all input and output neurons it connects (across all spatial positions).
If the resulting average is positive, the corresponding connection increases as it connects more simultaneously active neurons than neurons that fire disjointly. Conversely, if the average correlation is negative, indicating more disjoint firing, the connection strength is reduced.
Importantly, the weight update is applied only once per sample, after the full sequence of $T$ timesteps has been processed. Learning proceeds over multiple epochs, meaning each sample is revisited multiple times. Due to the small learning rate of $\eta=0.2$, these weight changes accumulate gradually.

\section{Experimental Results}\label{sec:exp_results}

\subsection{Experimental Setup}\label{sec:exp_setting}
Our experiments are designed to rigorously probe the \emph{core representational capability} of the CNA: its ability to form net fragments during training and recombine them into larger nets during inference. To isolate this property, we deliberately avoid task-specific or dataset-driven complexity and instead focus on the model’s behavior in its most fundamental setting.

We therefore train CNA on binary images of straight lines at varying orientations. Straight lines serve as the most elementary visual primitives: they are simple enough to avoid confounds yet naturally compose into richer structures such as line drawings of objects. This controlled regime directly tests whether CNA can learn fragmentary representations and later reassemble them into unseen global patterns. By contrast, benchmarks such as MNIST \cite{lecun1998gradient}, CIFAR \cite{krizhevsky2009cifar}, or ImageNet \cite{Deng_2009} involve dense, texture-heavy images where the link between local features and global structure is opaque. Evaluating CNA on such data at this stage would obscure its core representational mechanism, and, as we discuss in Section~\ref{sec:discussion}, would require introducing an additional object-representing area and more advanced filters, which we defer to future work.

Evaluation targets two key properties:
\begin{itemize}
    \item \textbf{Compositionality:} Can learned line fragments recombine into novel structures, such as object line drawings unseen during training?
    \item \textbf{Robustness:} Can nets be retrieved reliably under noise or occlusion?
\end{itemize}
To this end, we present the model \emph{after training and without updating any parameters} with (i) line drawings of objects composed of straight-line segments, (ii) inputs corrupted with Gaussian noise, and (iii) partially occluded patterns.

For comparison, we also train a conventional autoencoder on the same line-stimuli dataset to highlight representational differences. Further experimental details are provided in the appendix: We describe the training and testing datasets in more detail in Appendix \ref{sec:dataset}, the feature extractor mechanism to obtain $\boldsymbol{y}^{(S1)}$ in Appendix \ref{sec:features}, the training details of our CNA model in Appendix \ref{sec:parameters}, and the training details of the autoencoder in Appendix \ref{sec:autoencoder_params}.

\subsection{Interpreting Figures: General Considerations}

\begin{figure}[ht!]
  \begin{center}
    \includegraphics[width=\textwidth]{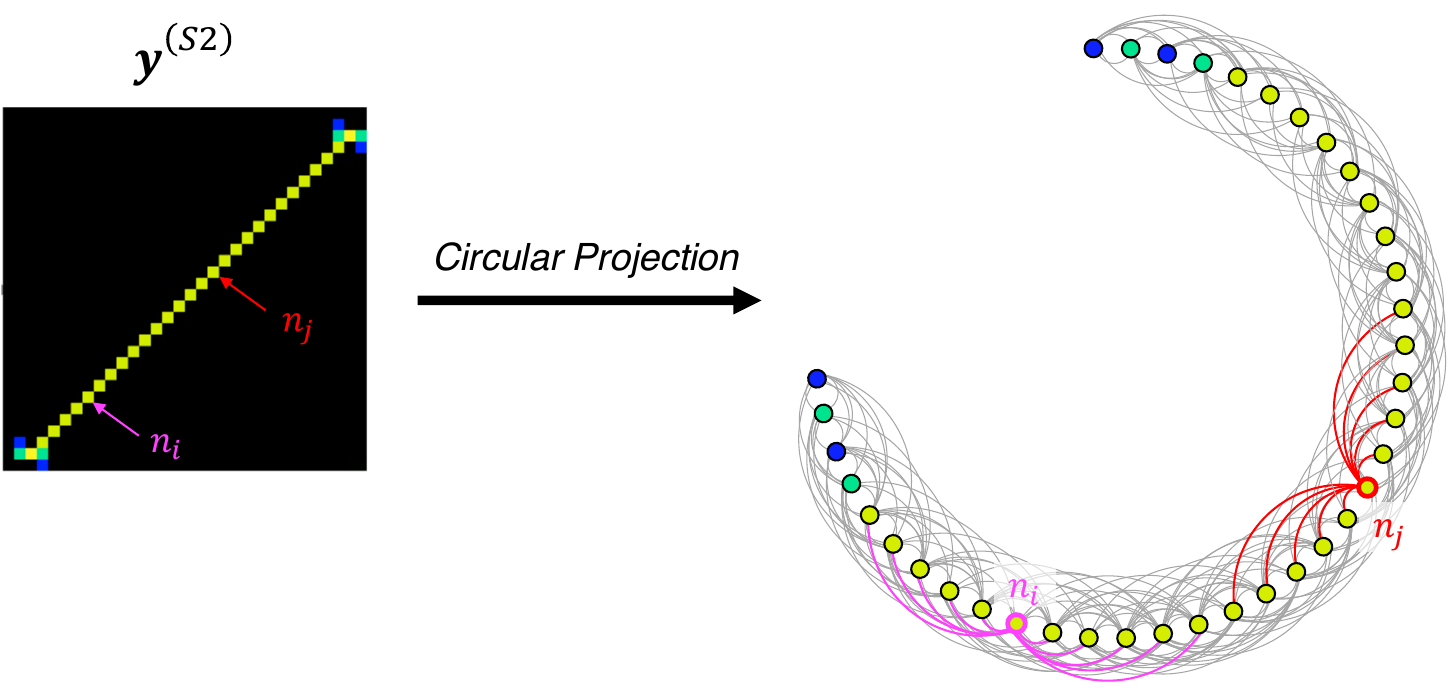}
  \end{center}
  \caption{%
  \textbf{The response of our model to a diagonal line input} is displayed as a net of activated neurons. The left of the image shows the neuronal activation $\boldsymbol{y}^{(S2)}$ when observing the line, and the right visualizes the net formed by these neurons (activations $\boldsymbol{y}^{(S2)}$ with their connections $\boldsymbol{W}^{(L)}$). For convenience, the neurons are rearranged in a circle, and the net fragment (connections) of two randomly selected neurons, $n_i$ and $n_j$, are visualized in colors (purple and red).}\label{fig:result_interpretation}
\end{figure}

A defining feature of CNA is that it represents patterns not through individual neuron activations, but through \emph{nets} -- cohesive, interconnected clusters of mutually supportive neurons. This network-based organization means that an individual neuron's activation is meaningful only in the context of its embedding within a larger net. Understanding CNAs' outputs, therefore, requires a shift in perspective: from isolated neuron activity to the connectivity and structure of the nets they form.

Figure \ref{fig:result_interpretation} illustrates this principle. After presenting a diagonal line input, the left panel shows the resulting neuronal activity $\boldsymbol{y}^{(S2)}$, while the right panel visualizes the underlying net structure derived from recurrent connections $\boldsymbol{W}^{(L)}$. For clarity, neurons are arranged in a circular layout, and the net fragments $\mathcal{F}(n_i)$ and $\mathcal{F}(n_j)$ of two randomly selected neurons $n_i$ and $n_j$ are highlighted in color. 

This visualization demonstrates two important points. First, individual neurons (e.g., $n_i$ and $n_j$) remain active only if embedded within supportive nets: without sufficient recurrent support, they are suppressed by attenuation mechanisms and would appear inactive (black) in the left panel. Second, this shows that observed activity patterns should always be interpreted as the visible footprint of underlying nets, even when those nets are not explicitly drawn.

Consequently, all subsequent visualizations of neural activity must be understood in this net-based context: the squares represent neurons that survive attenuation precisely because they are supported by their respective nets.

All our visualizations of neural activity are color-coded according to the competing channel groups $\mathcal{C}(n_i)$. Each color corresponds to a feature type (e.g., horizontal, vertical, or diagonal lines; see Appendix~\ref{sec:features}). However, these colors do not distinguish between different competitive channels $k \in {1, \dots, \kappa }$ within the same feature group, meaning that two neurons with the same color may belong to different competitive subnetworks with distinct connectivity patterns.

\subsection{Compositionality}\label{sec:results_compositionality}
\begin{figure}[ht!]
  \begin{center}
    \includegraphics[width=\textwidth]{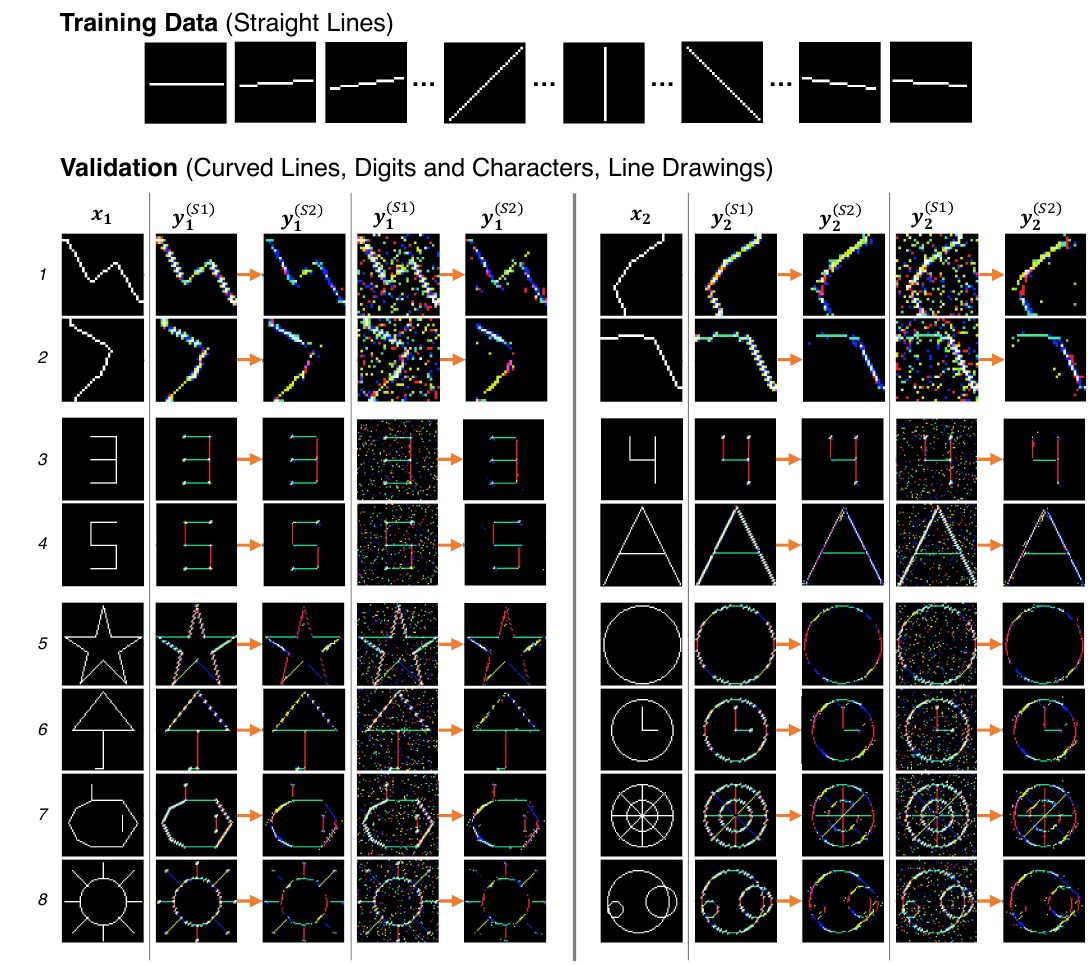}
  \end{center}
  \caption{%
  \textbf{Training and evaluation data.} The top of the figure illustrates the straight lines used during the training phase. Below, various samples from the evaluation datasets are shown, including kinked lines (rows $1$ and $2$), digits and characters (rows $3$ and $4$), and line drawings (rows $5$--$8$). For each row, two input examples, denoted as $\boldsymbol{x}_1$ and $\boldsymbol{x}_2$, are displayed alongside their respective feature activations, $\boldsymbol{y}^{(S1)}$, and neuronal activations, $\boldsymbol{y}^{(S2)}$, shown both without and with addition of Gaussian noise.
  }\label{fig:compositionality}
\end{figure}
The model weights are trained using straight lines (see training data examples at the top of Figure \ref{fig:compositionality}). During inference -- without additional learning -- the learned net fragments are recursively composed to form nets representing significantly more complex structures (see validation data in rows $1$--$8$ of Figure \ref{fig:compositionality}).
Notably, although such structures are absent from the training set, the CNA successfully generalizes to represent kinked lines (rows $1$ and $2$), digits and characters (rows $3$ and $4$), and line drawings (rows $5$--$8$).

These results demonstrate that while individual net fragments  $\mathcal{F}_t$ must be learned (in our case we use Hebbian learning to train the connections within $\mathcal{F}_t$), the recursive composition described in Eq. \ref{eq:recursive_net_fragments} to form $\mathcal{N}^*_t$ allows the integration of previously uncombined fragments. Consequently, the model exhibits combinatorial generalization through the structural recomposition of learned primitives.

\subsection{Filtering Gaussian Noise}\label{sec:results_noise_filtering}

The motivation for why we construct net fragments rather than just using the feature activation $\boldsymbol{y}^{(S1)}$ is demonstrated by introducing additional clutter:
For each sample, we generate an input representation both with and without Gaussian noise. This noise corresponds to invalid patterns not observed during training (i.e., are not captured in net fragments) that consequently lack sufficient recurrent support, leading to the deactivation of the corresponding neurons.
Thus, the CNA model only allows valid sub-patterns to persist and be composed into novel structures, while invalid patterns, such as noise, are suppressed.

\begin{figure}[ht!]
  \begin{center}
    \includegraphics[width=\textwidth]{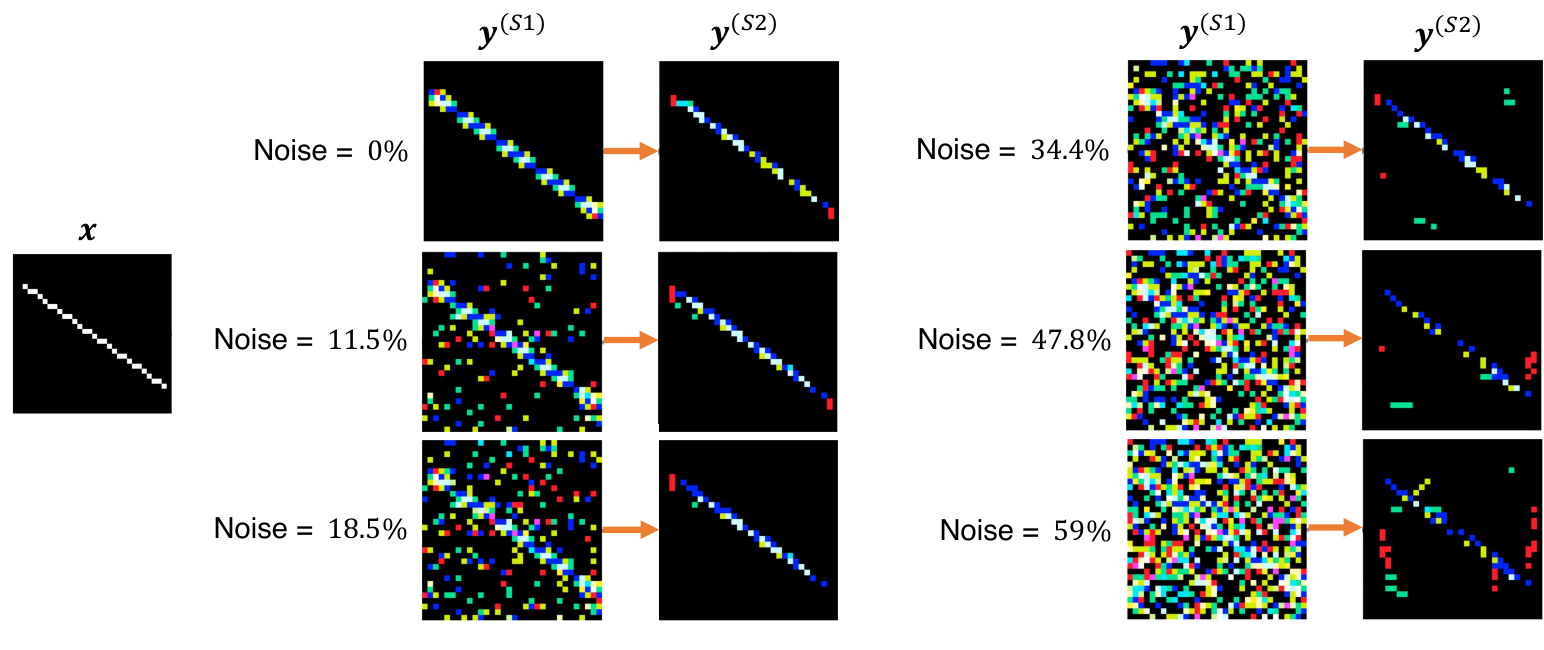}
  \end{center}
  \caption{\textbf{Noise filtering efficacy}: Noise-corrupted feature activations $\boldsymbol{y}^{(S1)}$ alongside the corresponding output $\boldsymbol{y}^{(S2)}$, in which noise is reduced due to a lack of support within nets. 
  }\label{fig:noise_filtering}
\end{figure}
To further investigate noise filtering, we introduce Gaussian noise of varying degrees (from $0\%$ to $59\%$ noise at each spatial location - resulting from adding up to $20\%$ noise per channel across four independent channels; see Appendix \ref{sec:dataset}) to the afferent input in stage $S2$, thereby simulating an image structure that has already survived the filtering of feature kernels in $S1$ (thus increasing the difficulty, as the fragments must handle all the noise rather than relying on the feature extraction stage).

The results are visualized in Figure \ref{fig:noise_filtering}: While introducing up to $47.8\%$ noise, remarkably consistent outputs $\boldsymbol{y}^{(S2)}$ are generated. As additional noise is introduced, more undesired neurons receive support and persist in their activity. However, the overarching pattern remains distinguishable even when subjected to noise levels of up to $59\%$.
In Appendix \ref{sec:quant_noise_filtering}, we quantitatively confirm the efficacy of noise filtering by measuring precision, recall, and noise filtration rate for various parameter settings, showing that lines remain well distinguishable for up to $59\%$ added noise.

\begin{figure}[ht!]
  \begin{center}
    \includegraphics[width=\textwidth]{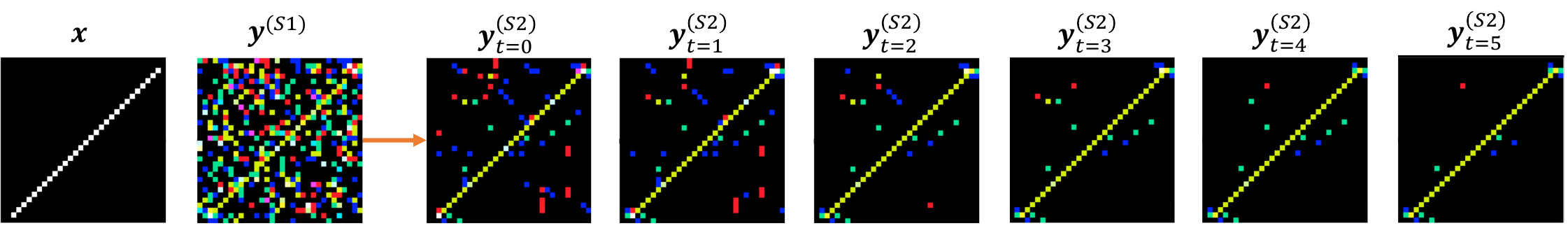}
  \end{center}
  \caption{\textbf{The filtering of (simulated) noise over time}: Already at time step $t=0$, most of the noise is suppressed. However, when attenuation increases over time, even more noise is filtered out. The plot is created after adding $10\%$ noise to each feature channel (using a bias $b^{(S2)} = 0.7$).
  }\label{fig:noise_time}
\end{figure}

The reason that the filtering works is that the additionally introduced clutter doesn't activate overlapping net fragments and cannot persist over time.
Figure \ref{fig:noise_time} depicts this systematic filtering across timesteps.
The figure illustrates a substantial noise reduction already at time step $t=0$ due to initial attenuation. In each subsequent time step, additional clutter is removed. This reduction is due to increased attenuation and falling support as a consequence of the drop-out of supporting neurons.

\subsection{Reconstructing Figures}\label{sec:partial_occlusion}

\begin{figure}[ht!]
  \begin{center}
    \includegraphics[width=0.95\textwidth]{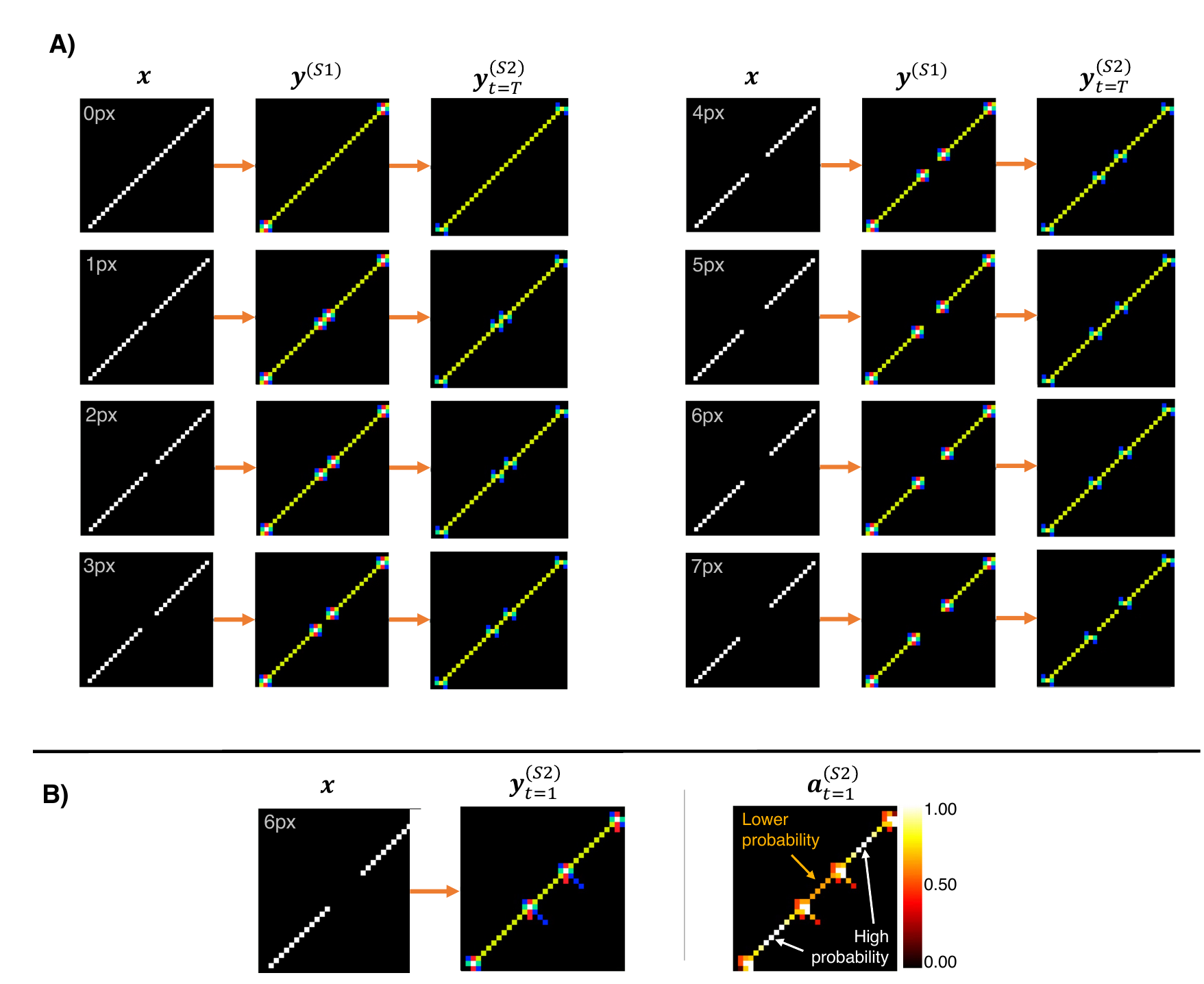}
  \end{center}
  \caption{%
  \textbf{Recovery from occlusion.} 
  Part A: %
  Input images of diagonal lines with different numbers of missing pixels, corresponding feature activations (middle column), and final output neuron activations. Activation bias $b^{(S2)}=0.5$, attenuation coefficient $\gamma=0.6+0.2t$. %
  Part B: %
  A diagonal line with $6$ missing pixels in the center (shown on the left), the corresponding output activity (in the middle), and the pre-synaptic activity $\boldsymbol{a}^{(S2)}$ displaying the ``activation probabilities'' (on the right) after the first time step.}
  \label{fig:li_examples}
\end{figure}
If a net fragment $\mathcal{F}_t(n_i)$ contains a substantial number of (active) neurons, an inactive neuron $n_i$ can receive significant support, leading it to switch on and encouraging figure reconstruction ($\mathcal{F}_t(n_i) \rightarrow \mathcal{F}^*_t(n_i)$).
Figure \ref{fig:li_examples}A shows how inactive neurons activate and thus demonstrates that nets can deal with occluded patterns.
This reconstruction is reliable for up to $3$ missing pixels (see Appendix \ref{sec:quant_noise_filtering}).
In some cases, as in this figure, the line is fully reconstructed for up to $6$ removed pixels and partly reconstructed for $7$ pixels.

\paragraph{Pre-Synaptic Activity Represents Expected Feature Activation.}
In the presence of missing pattern elements, the pre-synaptic activity map $\boldsymbol{a}^{(S2)}$ is higher at spatial locations where line features are observed and lower where they are not observed but expected (as shown in Figure \ref{fig:li_examples}B).
Depending on the context, it may be desirable either to reconstruct the missing line segment -- effectively bridging the gap between discontinuous features -- or to preserve the separation between distinct line elements.
This behavior can be modulated by adjusting the hyperparameter $b^{(S2)}$ in Eq. \ref{eq:s2_final_output}, which can govern the balance between feature completion and preservation of discontinuity, thereby allowing the model to selectively favor line reconstruction or separation. 
In future work, incorporating additional areas that maintain object-associated representations and act as top-down hypotheses could enable this selectivity to be learned rather than manually tuned.

\subsection{Comparison to Autoencoders}\label{sec:results_early_commitment}
\begin{figure}[ht!]
  \begin{center}
    \includegraphics[width=\textwidth]{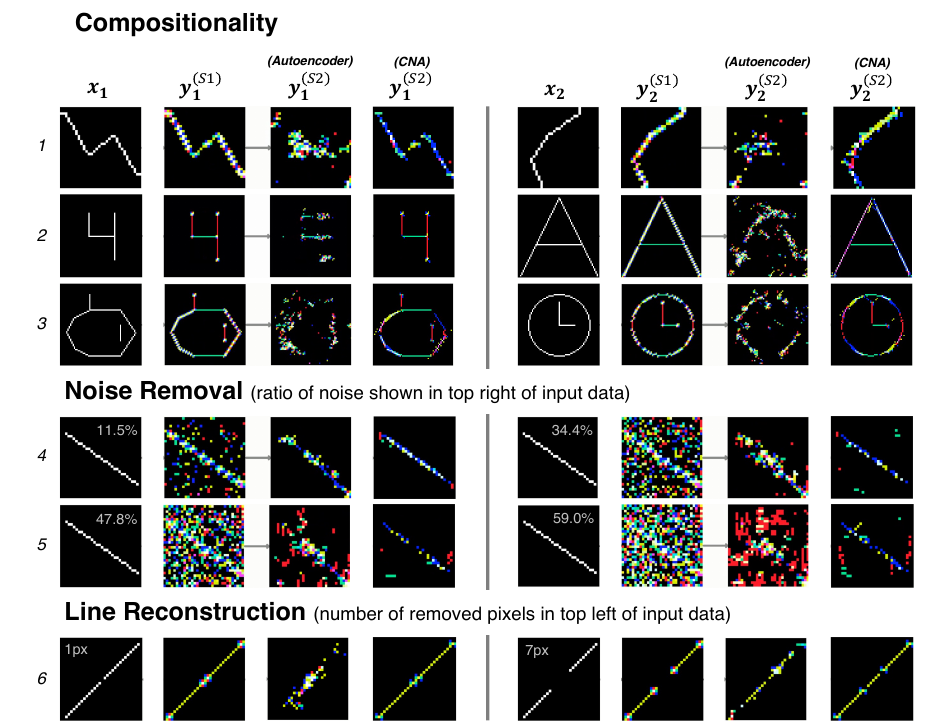}
  \end{center}
  \caption{\textbf{Comparison of the outputs generated by the proposed CNA model and an autoencoder}. Each row presents two samples, $\boldsymbol{x}_1$ and $\boldsymbol{x}_2$, alongside their respective feature activations and resulting representations (once for the CNA model and once for the autoencoder). Rows $1$--$3$ illustrate how both models encode patterns that were not present in the training dataset. Rows $4$--$5$ let one compare the models' performance in denoising tasks, while row $6$ displays their ability to reconstruct lines.}
  \label{fig:results_early_commitment}
\end{figure}

Figure \ref{fig:results_early_commitment}  compares performance between the proposed CNA model and an autoencoder (for model and training details, refer to Appendix \ref{sec:autoencoder_params}). While the autoencoder successfully learns to accurately represent line structures observed in the training data -- achieving a mean square error (MSE) of less than $1 \times 10^{-3}$ between the input and the reconstruction by the decoder -- it struggles with generalization to data not observed during training.
For instance, rows $1$--$3$ demonstrate that the CNA model is capable of representing previously unseen structures, while, in contrast, the autoencoder fails to capture such out-of-distribution patterns.

Rows $4$--$5$ highlight the autoencoder's ability to represent line structures with up to 11.5\% noise introduced. However, as the noise level increases, its performance decreases significantly. The CNA model, in contrast, remains robust under much higher noise levels, tolerating up to $59\%$ noise while still producing high-quality representations. Metrics on noise reduction, precision, and recall, presented in Appendix \ref{sec:quant_noise_filtering}, further substantiate the CNA model's superior performance.

In row $6$, the models' capabilities for reconstructing missing line segments are compared. Both models can reconstruct the removed pixels, but the autoencoder introduces additional noise in regions other than the line center, where the pixels were removed.
This observation is confirmed by the metrics in Appendix \ref{sec:quant_noise_filtering}: The autoencoder is slightly superior in reconstructing the removed pixels, but this comes at the cost of increased noise elsewhere, leading to lower overall precision and recall for figure reconstruction.

\section{Relation to Alternative Concepts} \label{sec:RelatedWork} 
The central issue behind our model is how the structure of the environment is represented. In biology, this structure would be accessed by the senses in the form of sensory patterns --- large active subsets of sensory elements (e.g., retinal ganglion cells). While global sensory patterns vary across different conditions and viewpoints, local sensory structure tends to exhibit consistency. For instance, the sensory pattern associated with a static object may appear different under varying eye positions or perspectives, yet local fragments of the pattern remain stable and coherent (we demonstrate this in our experiments by having constant straight-line patterns that are stabilized, while changing Gaussian noise is ignored).
The task of learning -- both in biology and in a model like ours -- is to pick up durable structures from sensory patterns in spite of their fleeting nature. We proceed here with a simple assumption regarding this durable structure: There is an important class of coherent sensory patterns that are composed of a finite set of overlapping pattern fragments. We assume that these pattern fragments can be represented by the recurrent connectivity of net fragments and that this connectivity can be learned from the statistics of sensory patterns with the help of Hebbian plasticity.  This is the background against which we discuss some models of neural computation currently dominating the literature.

\subsection{Associative Memory} \label{sec:associative_memory}

The central concept in the classic works of \cite{hebb_organization_1949}  and \cite{hayek} is the assembly, a set of neurons supporting each other's activity by mutual excitatory connections, connections that have been created by synaptic, ``Hebbian'', plasticity.  This concept has found a concrete mathematical formulation as ``associative memory'' (AM) \citep{amari_learning_1972, Hopfield1982}, where stored patterns are encoded by enhancing excitatory connections between all pairs of co-active neurons.

While AM mechanisms are effective for storing and retrieving known patterns, they often fail when applied to perceptual tasks such as object recognition, where previously unseen structures must be classified based on prior experience.
This limitation arises because AM tends to store complete patterns as static ``snapshots.'' 
The reason for this rigidity of AM states is their mode of recording: The excitatory connections between all pairs of neurons active at a particular moment are strengthened simultaneously, without regard to whether a connection is significant in the sense of applying to many patterns (that is, being part of a repeating pattern fragment) or insignificant, applying only to this particular pattern.
  
In our model, in contrast, the synaptic weights are gradually modified in response to the statistical properties of a large ensemble of sensory patterns. According to our Eq. \ref{eq:hebbian_plasticity}, a particular synapse can only grow if the pair of neurons it connects is simultaneously active again and again (indicated by positive correlation $\boldsymbol{\rho}_{avg}$), or it decays due to negative correlation.
This processing allows the system to gradually extract and store re-occurring (statistically relevant) sub-patterns rather than memorizing isolated global patterns.

A functional restriction on AM is that different stored states have to be statistically independent, that is, as sets of active neurons, they have to have low pairwise overlap. This is an awkward constraint, as sensory patterns are composed of repeating pattern fragments and thus are bound to have a large overlap. In classical AM, this constraint has been addressed just by requiring patterns to be small subsets of all neurons \citep{Tsodyks2007, Palm2013}. Our model can address the small-overlap constraint in a systematic way with the help of multiple neurons (``competing neurons'') that respond to the same trigger feature but compete with each other for activity so that pairs of pattern fragments that have high feature-wise overlap can select different competing neurons, resulting in the corresponding net fragments having low neuron-wise overlap. This is made possible by the statistical mode of strengthening synapses and is not possible in AM with its one-shot learning mode.

Our model also addresses an open issue of AM: The selection of coherent sensory patterns for storage. As coherence is defined as being composed of a fixed set of overlapping pattern fragments (stored in net fragments), it is only natural to require that the learning process converges to a connectivity state in which only sets of neurons that segment the input into coherent sensory patterns can be activated stably \citep{Malsburg2024}.

The canonical formulation \citep{amari_learning_1972,Hopfield1982} of the AM's state dynamic defines an ``energy'' that acts as a Lyapunov function. Beginning from any initial state, neurons are visited sequentially, and their state is flipped to gradually reduce the energy (simultaneous switching of the neurons would lead to non-sensical oscillations). Our model's dynamics is not based on a Lyapunov function and a time-consuming gradual process, relying instead on an externally imposed periodic attenuation control parameter that lets the set of neurons originally activated by the sensory input collapse to a subset formed by neurons that support each other by excitatory connections.

\subsection{Boltzmann Machine} \label{boltzmann_machine}

Regarding Boltzmann Machines (BM), we refer to \citet{Hinton1983, hinton_constraint1984}, taking the latter as the authoritative reference. Both the BM and our model describe networks of recurrent connections between neurons, where some neurons get activated by sensory input (``visible units''), and by employing synaptic plasticity, both approaches aim to turn the system into a constraint satisfaction network that uses inner connections to stabilize the patterns suggested by the sensory input.
  
The BM as described in \citet{hinton_constraint1984} and our model differ, however, in a number of ways.  Inspired by statistical mechanics, the BM speaks of equilibrium macro-states, each of which is realized as an ensemble of micro-states, the probability of a micro-state being proportional to a negative exponential of a state ``energy'' (Boltzmann distribution), which is defined as in Amari-Hopfield's Associative Memory as a function of neural activities and connection weights. The BM further assumes that the system compares two modes, clamped (in which the sensory neurons are determined by the input) and free running (no sensory input), and lets synaptic plasticity equalize the signal correlations of pairs of neurons observed in one of the two modes with those observed in the other.
In our model, neural activity is not modeled in terms of probabilistic ensembles of states but
as discrete-time trajectories over a fixed number of time steps $T$, during which the same sensory input is constantly refined.
In distinction to the Boltzmann Machine, these trajectories evolve deterministically, according to update rules that depend on both the input and recurrent connectivity.
Importantly, the dynamics are externally modulated: At each step, a control mechanism attenuates the activity of neurons that lack sufficient recurrent support from other active neurons;
see Eqs. \ref{eq:inhibition} and \ref{eq:s2_final_output}.
In contrast, Boltzmann Machines rely on stochastic sampling over ensembles of network states, governed by a distribution derived from an energy function. While the statistical mechanics framework underlying Boltzmann Machines is elegant and conceptually rich, it is computationally inefficient in settings where a large number of intermediate states must be generated and compared to approximate equilibrium. Our model avoids this bottleneck by converging rapidly through deterministic dynamics over a small fixed number of steps.

A necessary precondition for the feasibility of the goals of the BM and our model is the existence of a system of constraints on the sensory patterns that can be learned from a finite number of such patterns and which then suffice to stabilize all future patterns.  Given the bias-variance dilemma \citep{Geman1992} and the no-free-lunch theorems \citep{NoFreeLunch}, a learning system's architecture needs to drastically constrain its search space by a bias that is tuned to the application domain \citep{learning1997tom}. No such bias is built into the BM's architecture in its generality \footnote{The concrete application described in \citet{hinton_constraint1984} (the `Encoder Problem') provides such bias, but is trivial in the sense of explicitly defining the full (and very small) set of sensory patterns.}, while our model's architecture explicitly incorporates the bias that the sensory patterns of its application domain, vision, are all \emph{coherent composites of a finite set of pattern fragments}, a bias that is effective yet captures an essentially infinite number of sensory patterns.

The realization that the BM in its generality would imply unfeasible learning times led to the introduction of Restricted Boltzmann Machines (RBM) \citep{Smolensky1987,Hinton2006}. RBMs consist of two layers, one visible and one latent, with bidirectional connections between them, but no connections within them. A stack of RBMs is used in \citep{Hinton2006}
to construct an autoencoder in which RBM learning is used to pre-train the connections between layers in a sequence from bottom (input) to top (the sparse coding layer). Then, the whole encoder-decoder stack is trained end-to-end by backpropagation of error.
We discuss the differences between our model and networks trained by backpropagation of error (including RBMs) in the next sub-section.

\subsection{Artificial Neural Networks (ANN)} \label{sec:ANN}

Artificial neural networks (ANNs), as concerned here, are layered structures with feedforward layer-to-layer connections from input to output, and these connections are trained by back-propagation of error \citep[p 292]{rosenblatt_principles_1962}. Prominent examples are multilayer perceptrons (MLPs) \citep{prince2023understanding}, CNNs \citep{lecun1998gradient}, autoencoders \citep{Hinton2006}, and transformers \citep{attention2017}.  In these systems, individual neurons in one layer represent
feature compositions 
that occur in the layer below.  The complexity and variability of features represented by individual neurons grow from layer to layer  (as shown in reconstructions, see \citet{Zeiler2013}). The hierarchy of layers thus constitutes a hierarchy of nested pattern fragments. 

The architectures of both ANNs and our model implement the bias that a virtually infinite set of potential input patterns can be represented in the form of flexible compositions of a finite set of pattern fragments.  They differ fundamentally, however, in how these pattern fragments are implemented, how they are combined, and how they are learned. 

Beginning with the latter, ANNs select the features to be represented on the basis of their effect on lowering the system's prediction error, whereas in our model the selection of features to be represented by net fragments is entirely a matter of the statistics of input patterns (as is, incidentally, the case with the RBM pretraining of layer connections in \citet{Hinton2006}). 

Regarding implementation, ANNs represent elementary pattern fragments in the first latent layer by feature neurons. In our model, which has only one layer, elementary pattern fragments are represented by minimal net fragments, each composed of a neuron and a number of excitatory connections from other neurons within the same layer. That a neuron remains active under attenuation indicates that it has recognized the pattern formed by these other also activated neurons. ANNs represent a nested hierarchy of pattern fragments, as higher-level neurons respond to combinations of active lower-level neurons.  Our model equally represents a nested hierarchy of pattern fragments within one layer, in this case composed of a neuron, its immediate supporters, the supporters of the supporters, and so on. 

The rules for composing larger structures out of smaller pattern fragments correspondingly differ between ANNs and our model. In CNNs, this flexibility is mostly implemented by max pooling, or in transformers by the attention mechanism that selects active neurons in one layer to contribute to the activity of neurons in the next. This freedom of choosing pattern fragments in lower levels is exerted independently by the neurons in a given level, without regard to the mutual compatibility of those selected lower-level patterns. This lack of coherence is even more drastic between ANNs' high-level feature neurons, which don't express in any way mutual overlap in terms of constituent sub-patterns (the binding problem, see \citet{Greff2020}), the likely cause of vulnerability of ANNs to adversarial attack \citep{Leadholm2022}. In our model, in contrast,  the net fragments that respond to different pattern fragments have to fit together into one consistent net, in which each active neuron is sufficiently supported by other active neurons. This expresses the bias that significant patterns in the visual environment are composed of pattern fragments {\it in a coherent way}. We consider this as the solution to the binding problem \citep{Greff2020} and as an explanation of the Gestalt phenomenon \citep{Wagemans2012}.

\section{Discussion and Outlook}\label{sec:discussion}
The Cooperative Network Architecture (CNA) introduced here proposes a novel approach to neural representation by constructing structured, composable nets from input patterns. 
It is based on the vision to overcome the tension that has eternally been plaguing the brain/mind issue and AI, the tension between low-level and high-level, between sub-symbolic and symbolic descriptions ({\it cf.} \cite{FodorPylyshyn}). Both perspectives have their strengths and their weaknesses (as discussed by \citet{Berlin2023}).
While our model preserves global, high-level pattern structure through the composition of net fragments (the symbolic aspect), it also retains sensitivity to local detail via the fine-grained connectivity of its elements (the sub-symbolic aspect). This synthesis of descriptive levels is the basis for perceptual segmentation (as in the Gestalt phenomenon), noise filtering, pattern completion, and for pattern-global decisions such as classification or recognition (see below).

Empirically, we demonstrated that CNA generalizes beyond its training distribution, successfully reconstructing unseen objects and filtering substantial noise. These findings validate CNA’s ability to form holistic, yet flexible, representations. Furthermore, the reuse and recombination of learned sub-patterns provide a strong inductive bias, allowing for compositional generalization from limited data.

Conceptually, our model deviates in essential ways from classical neural models. In contrast to the perceptron (which is currently dominating AI), the structure of patterns is expressed by recurrent and not feedforward connections, the entirety of patterns is expressed as structured nets instead of by structureless individual neurons, and learning of connectivity is based on Hebbian plasticity and input statistics instead of backpropagation of error and human-provided sample (and ground truth) material. In contrast to the associative memory model, patterns have a flexible articulated structure, being composed of generic pattern fragments, instead of monolithic rigidity. From the Boltzmann Machine, our model differs by being tuned to the natural environment by a powerful bias, which permits to capture a virtual infinity of patterns after learning from a finite set of examples.

What is the functional role of nets within intelligent systems?  According to a long-standing proposal, invariant object recognition can be realized by structure-sensitive, homeomorphic mappings between networks \citep{von_der_malsburg_correlation_1981, anderson_shifter_1987, Olshausen1995, wiskott_face_1997, Arathorn2002, wolfrum_recurrent_2008, Memisevic2010, von_der_malsburg_theory_2022}. This calls for the existence of a pair or a series of layers, the lowest of which is the sensory layer we have modeled here, while higher layers harbor nets of growing invariance or generality.  Nets in higher layers can be invariant against transformations such as translation, rotation, or scaling in the image plane, or can be more abstract by leaving out submodalities such as color or texture, or by accommodating distortion. Dynamic mappings that can keep up with rapid changes within the input layer are realized in the cited literature by the rapid selection of connections that connect neurons of equal feature type and that preserve the neighborhood relationships within the connected nets. Homeomorphic mappings can themselves be considered as nets of a second kind, and they also are coherently composed of overlapping map fragments \citep{anderson_shifter_1987, ZHU2004}.  Several nets tied together by mappings can be considered as nets on a higher level. The purpose of the model we have presented here is to fill a sore gap in this line of work in providing a clear picture of how nets that are eligible for mapping are dynamically formed.

Urgent future work will have to expand the present model to cope with a full range of natural visual input and a rich range of visual submodalities (texture, color, stereo depth, motion), will have to expand to other modalities (e.g., audition, motor control), and will have to combine the functional elements modeled here and in the cited work into fully functional systems.

\section*{Funding}
This research has been funded by the Canton of Zurich, Switzerland, through the Digitalization Initiative of the Canton of Zurich (DIZH) Fellowship project `Stability of self-organizing net fragments as inductive bias for next-generation deep learning.'

\section*{Data Availability Statement}
The code to generate the data as well as to reproduce the results can be found after the publication of this work on Github at \url{https://github.com/sagerpascal/cna}.

\appendix

\section*{Appendix}

\section{Feature Extraction}\label{sec:features}
\begin{figure}[ht!]
  \begin{center}
    \includegraphics[width=0.7\textwidth]{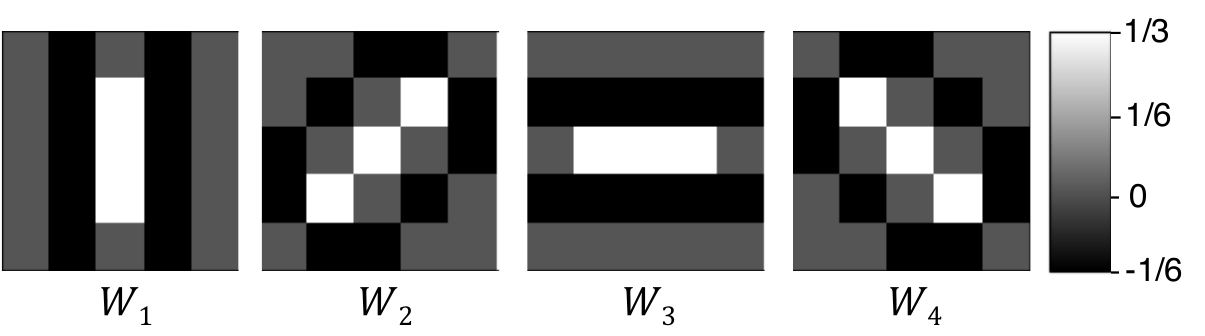}
  \end{center}
  \caption{\textbf{Features}: The hand-crafted convolutional kernels of the feature extractor.}\label{fig:weights_feature_extractor}
\end{figure}

The stage $S1$ serves to extract initial binary features from the images.
The input images have $C=1$ channel and $C^{(S1)}=4$ features are extracted at each spatial location. We utilize hand-crafted filters with a size of $5 \times 5$ as displayed in Figure \ref{fig:weights_feature_extractor}, each of the four filters corresponding to a particular line orientation (vertical, positive diagonal, horizontal, and negative diagonal).
This approach is motivated by the observation that straight lines when viewed locally, can be expressed as combinations of these fundamental line types (e.g., a line with a $20^{\circ}$ angle activates the horizontal and the $+45^{\circ}$ diagonal feature neurons). The use of such manually designed filters allows a more straightforward interpretation of the results. While more complex datasets necessitate learned filters, our primary focus in this work is on constructing net fragments, and thus, we favor the simplicity of hand-crafted filters.

These filters move across the image with a step size (`stride') of $1$ (assuming $0$ as input signal beyond the image border: `zero padding') so that the input image and output image are of the same size.
Thus, the input $\boldsymbol{x}$ in the feature extraction stage are images of size $(1\times32\times32)$ and the output $\boldsymbol{y}^{(S1)}$ feature activation maps of size $(4\times32\times32)$. The dataset's lines activate multiple filters at different positions, facilitating the construction of net fragments across these feature channels.
We convert the floating-point output of the convolutional weights to binary activations using the heaviside function $\Theta$ with a bias of $b^{(S1)}=0.5$, i.e. neurons fire a binary spike if $\boldsymbol{a}_j > 0.5$ (see eq. \ref{eq:y_s1}).

\section{Normalization}\label{sec:normalization}
The pre-synaptic activity in $S2$ is normalized using a two-step approach that confines the neuronal activations $\boldsymbol{a}^{\mathcal{I}(S2)}$ to the interval $(0,1)$. 

We first apply a size limit:
\begin{equation}
  \boldsymbol{a}^{\mathcal{I}(S2)}_{c,t} = \begin{cases}
   		\boldsymbol{a}^{\mathcal{I}(S2)}_{c,t}, & \text{ if } \boldsymbol{a}^{\mathcal{I}(S2)}_{c,t} \leq \lambda \\
   		\lambda - \frac{1}{2}(\boldsymbol{a}^{\mathcal{I}(S2)}_{c,t} - \lambda), & \text{otherwise}.
  	\end{cases}\label{eq:saturation}
\end{equation}
According to this formula, $\boldsymbol{a}^\mathcal{I}$ grows undiminished with $\boldsymbol{a}^\mathcal{I}$ until it reaches the value $\lambda$, and thereafter actually diminishes with slope $-1/2$. (In our experiments we found the value $\lambda = 1.3 \cdot \frac{h + w}{2}$ to work well). 
This saturation helps to mitigate imbalances between net fragments with differing numbers of participating neurons. In the next step, we introduce the normalization
\begin{equation}
    \boldsymbol{a}^{\text{norm }(S2)}_{c,i,t} = \max \left( \frac{\boldsymbol{a}^{\mathcal{I}(S2)}_{c,i,t}}{\max_{i'} \boldsymbol{a}^{\mathcal{I}(S2)}_{c,i',t}}
    \label{eq:normalization}, 0 \right)
\end{equation}
Here, $i$ refers to spatial locations within the image, i.e., $i=(h,w)$ where $h \in H, w \in W$.
This normalization confines activations to the interval $[0,1]$, where neurons receiving maximal support are mapped to $1$, while those with less support are mapped to correspondingly smaller values. 
This normalization is relative to the maximal value that itself is growing during learning as it is important to give neurons initially (when recurrent connections are still small) a chance to fire and later when recurrent weights between correlated patterns have been learned, to suppress those neurons that are part of noise patterns. 

\section{Single Weight-Kernel Implementation}
So far, we introduced the weights $\boldsymbol{W}^{(F)}$ and $\boldsymbol{W}^{(L)}$ in $S2$ as two independent convolutional kernels.
For simplicity in our algorithm, and since we use the same kernel size for the forward and recurrent connections, we stack the two kernels as $\boldsymbol{W}^{(S2)} = (\boldsymbol{W}^{(F)}; \boldsymbol{W}^{(L)})$ as well as the inputs $\boldsymbol{y}^{(S1)}$ and $\boldsymbol{y}^{(S2)}_{t-1}$, leading to the equivalent formulation of eq. \ref{eq:a_a2}:

\begin{equation}
  \boldsymbol{a}^{\mathcal{I}(S2)}_{c_{\text{out}},t} = \sum_{c_\text{in}=0}^{C^{(S2)}} \boldsymbol{W}^{(S2)}_{c_{\text{out}}, c_{\text{in}}} \star \left( \boldsymbol{y}_{c_{\text{in}}}^{(S1)}; \boldsymbol{y}_{c_{\text{in}},t-1}^{(S2)} \right)
  \label{eq:a_a22}
\end{equation}

\section{Dataset} \label{sec:dataset}
The dataset used for training comprises binary images measuring ($32 \times 32$) pixels, each sample depicting a straight line going through the image center, starting and ending $2$ pixels from the image boundary  (leading to $59$ distinct lines of different angles).
The images are generated when required and may vary from one epoch to another.
During each training cycle, we randomly generate $300$ image instances (each line about $5$ times).
During evaluation, we sample each of the $59$ distinct lines exactly once.
In contrast to training, the samples used for evaluation comprise local distortions in the form of additive Gaussian nose and missing line segments (subtractive noise), and we evaluate the net fragments' capability to remove these.

To introduce additive Gaussian noise to the afferent input in stage $S2$, we probabilistically flip neurons within each channel with a probability of up to $20\%$. Since we use four feature channels in our experiments, this corresponds to a probability of $1-(1-0.2)^4=59\%$ that a neuron is flipped at any spatial location. 
To evaluate subtractive noise, we create discontinuous lines by deleting a line segment of $1$ to $7$ pixels in the middle.

To demonstrate the generalization capability of CNA, we generate validation samples that differ from the ones used during training. Specifically, we produce binary images measuring ($32 \times 32$) pixels depicting kinked lines by generating four random points within the image and drawing straight lines between these points.

Additionally, we generate binary images measuring ($64 \times 64$) pixels by hand that represent more complex structures.
These images include line drawings of digits $0$-$9$, characters, and objects such as a house, a star, a wheel, a clock, etc. Some of these images are shown in Figure \ref{fig:compositionality}, and an extensive list can be found at \url{https://github.com/sagerpascal/cna}.

\section{Training Parameters}\label{sec:parameters}
We train the model with a learning rate of $\eta=0.2$ for $100$ training cycles (epochs). The model uses $\kappa=10$ competitive neurons with recurrent connections spanning a kernel of $11 \times 11$ neurons.
Since we use $C^{(S1)}=4$ filters for feature extraction in $S1$ (see Appendix \ref{sec:features}), the number of competitive channels in $S2$ corresponds to $C^{(S2)} = 4 \cdot \kappa = 40$.
The resulting weight matrices have dimensions $\boldsymbol{W}^{(F)} \in \mathbb{R}^{40 \times 4 \times 11 \times 11}$ and $\boldsymbol{W}^{(L)} \in \mathbb{R}^{40 \times 40 \times 11 \times 11}$, respectively $\boldsymbol{W}^{(S2)} \in \mathbb{R}^{40 \times 44 \times 11 \times 11}$.

\section{Autoencoder Training}\label{sec:autoencoder_params}
We compare our CNA model to an ANN, specifically an autoencoder \citep{Hinton2006}, which is a widely used self-supervised technique for generating input representations. An autoencoder is particularly suited for comparison because it includes a built-in decoder, allowing for feature reconstruction and visualization, which is required for feature comparison. To ensure a fair comparison, we also employ convolutional filters \citep{lecun_backpropagation_1989} to the autoencoder and incorporate state-of-the-art design principles (we even permit more data and feature channels to the autoencoder).

The autoencoder is trained to reconstruct the feature activations, $\boldsymbol{y}^{(S1)}$, using the same training dataset as our CNA model. It consists of an 8-layer architecture, with an encoder composed of convolutional layers with a kernel size of $3$ and a stride of $2$, with progressively increasing channel counts of $32$, $64$, $128$, and $256$ channels. The decoder mirrors this structure, using transposed convolutional layers with $256$, $128$, $64$, and $32$ channels, respectively. ReLU activations are applied after each layer. Consequently, the autoencoder is a relatively large model with $777$k parameters, utilizing up to $256$ channels, whereas the CNA model operates with only $44$ channels.

The autoencoder is trained by minimizing the L2 norm between the input features and their reconstructed counterparts, using the Adam optimizer \citep{Kingma_Ba_2015} with a learning rate of $1 \times 10^{-4}$, beta parameters of $(0.9, 0.999)$, and a weight decay of $1 \times 10^{-8}$.

To provide a robust comparison, we increase the number of training samples for the autoencoder. Rather than training for 100 epochs with 300 samples per epoch as done for the CNA model, we train the autoencoder for $200$ epochs, using $10,000$ images per epoch. This results in a significantly larger training set but ensures that the autoencoder achieves a highly accurate reconstruction. The autoencoder is trained with a batch size of $512$.

\section{Quantifying Noise Filtration}\label{sec:quant_noise_filtering}

Our evaluations focus on the efficacy of net fragments in suppressing added Gaussian noise and complementing missing or occluded patterns. For both evaluation experiments, we report {\it recall} and {\it precision}: Recall is defined as the proportion of neurons that maintain activity (on) when noise is added, and precision quantifies the extent to which neurons activated in the presence of noise were also active without noise.
To assess the system's noise reduction capability in the case of Gaussian noise, we additionally measure the percentage of flipped neurons that revert to their original activation state, referring to this measurement as the {\it noise reduction rate}.
To assess subtractive noise, we measure the similarity between the original (created without noise) and reconstructed (created with noise) feature activations at the spatial locations where pixels have been removed. We call this metric the {\it feature reconstruction rate}.

\paragraph{Filtering Gaussian Noise.}
\begin{figure}[ht!]
  \begin{center}
    \includegraphics[width=\textwidth]{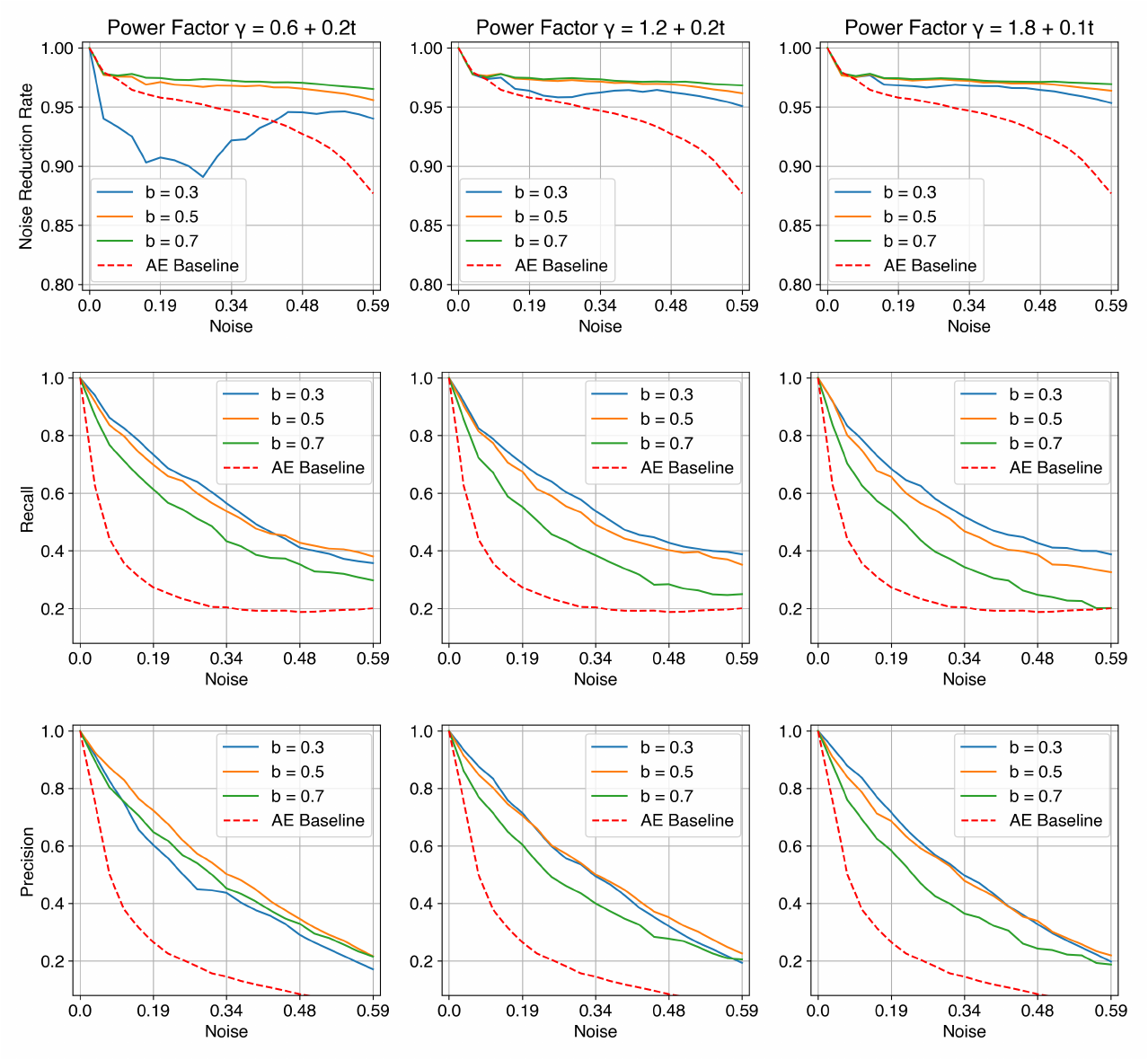}
  \end{center}
  \caption{\textbf{System performance as function of the noise level}. First row: noise reduction rate; second row: recall; third row: precision. Columns differ in the attenuation coefficient $\gamma$. The colors within each plot represent different activation biases $b^{(S2)}$. The dotted red line is the autoencoder, serving as baseline.}\label{fig:noise_results}
\end{figure}
Figure \ref{fig:noise_results} displays noise reduction rates, recall, and precision across different parameter configurations for different levels of added noise.
The metrics demonstrate that increasing the attenuation coefficient or the activation bias results in enhanced noise filtration.

The overall efficacy of noise reduction remains consistently high across most parameter configurations, exhibiting a constant noise reduction rate of $\geq \approx 95\%$. This rate persists regardless of the magnitude of the introduced noise. As the noise suppression rate is approximately constant, this means that if there is more noise in the input, there will also be more noise in the output, albeit limited to around 5\% of the input noise.

Furthermore, all CNA models (except the one with the lowest power factor and bias) outperform the autoencoder, demonstrating that they achieve performance that is competitive with state-of-the-art deep learning models.

\paragraph{Reconstructing Subtractive Noise.}
\begin{figure}[ht!]
  \begin{center}
    \includegraphics[width=\textwidth]{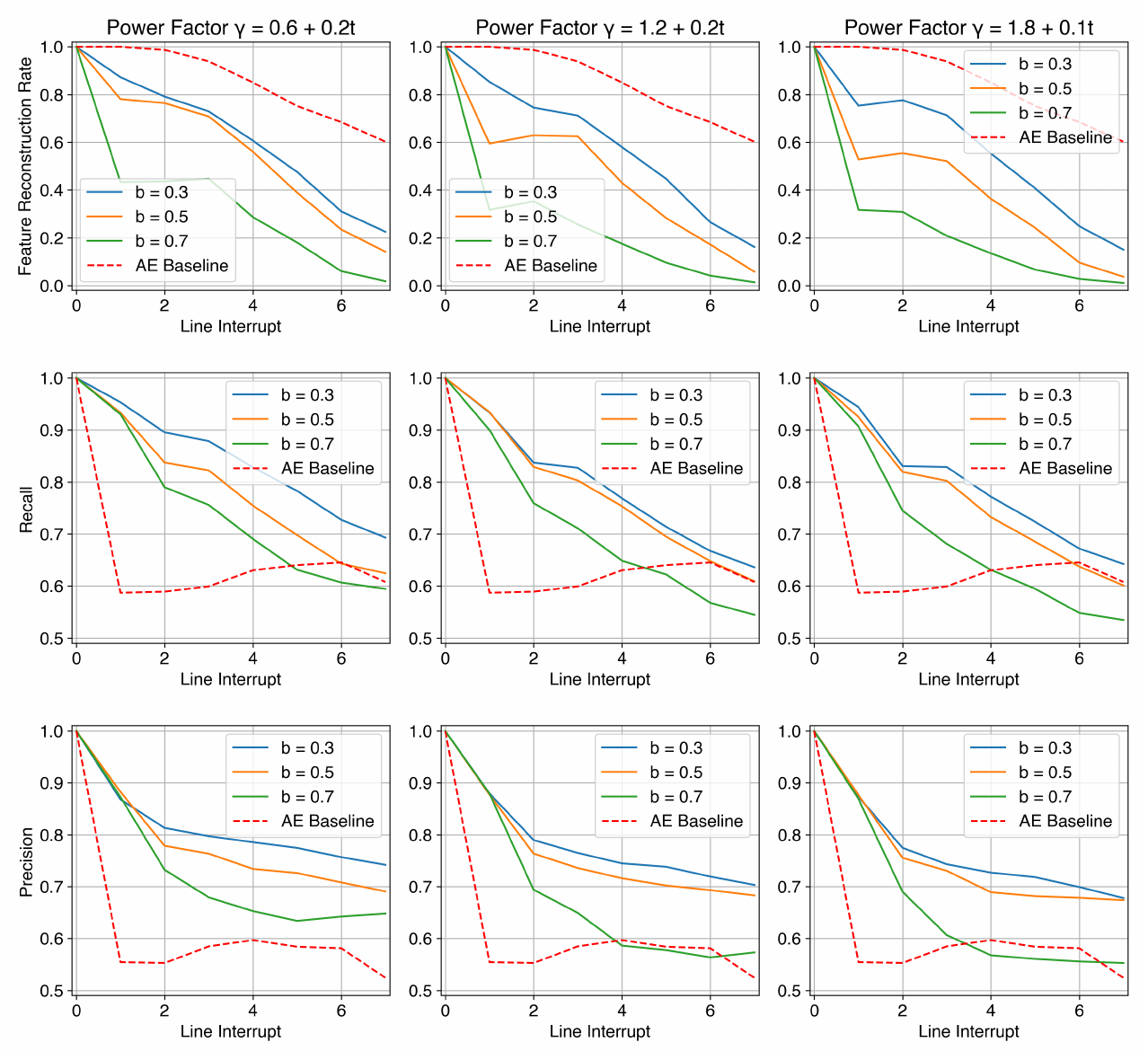}
  \end{center}
  \caption{\textbf{Reconstruction of partial occlusion}. {\it First row:} feature reconstruction rate; {\it second row:} recall; {\it third row:} precision. Columns are distinguished by the attenuation coefficient $\gamma$, and line colors within plots distinguish activation bias $b^{(S2)}$. The dotted red line is the autoencoder model.}\label{fig:li_results}
\end{figure}

Figure \ref{fig:li_results} displays the feature reconstruction rate, precision, and recall for removed line fragments of different lengths. Having a lower activation bias or attenuation coefficient $\gamma$ leads to a better reconstruction of removed patterns, as these settings allow neurons with less support to activate.

Compared to the autoencoder baseline, the CNA model achieves higher recall and precision but a lower feature reconstruction rate.
This indicates that the autoencoder is superior at reconstructing the pixels that have been removed but that, in general, much more noise (other neurons are activated/deactivated) is introduced.

\paragraph{Interpreting Metrics.}
While the reported values for precision and recall may appear low, we argue that these fragments are of good quality and point out that precision and recall metrics have to be interpreted differently than in a classification context: 
A net fragment consisting of many neurons characterizes (a part of) an object and maintains this characterization even when many of its neurons are inversed (and precision and recall are low). This aligns with the findings of \citet{ahmad_properties_2015}, highlighting that binary distributed activations demonstrate high robustness, retaining accurate interpretation even in the presence of numerous flipped neurons.

In the case of the depicted straight lines, the feature neuron activations of the two most similar lines overlap with $21.2\%$. Therefore, achieving precision and recall values surpassing this threshold permits reliable discrimination even among similar lines, which is the case for the CNA but not for the autoencoder (e.g., see Figure \ref{fig:noise_results}).


\clearpage
\newpage

\begin{thebibliography}{}

\bibitem[\protect\astroncite{Ahmad and Hawkins}{2015}]{ahmad_properties_2015}
Ahmad, S. and Hawkins, J. (2015).
\newblock {Properties of sparse distributed representations and their application to hierarchical temporal memory}.
\newblock {\em arXiv}, 1503.07469.

\bibitem[\protect\astroncite{Amari}{1972}]{amari_learning_1972}
Amari, S.-I. (1972).
\newblock {Learning patterns and pattern sequences by self-organizing nets of threshold elements}.
\newblock {\em IEEE Transactions on Computers}, C-21(11):1197--1206.

\bibitem[\protect\astroncite{Amirian et~al.}{2018}]{amirian2018trace}
Amirian, M., Schwenker, F., and Stadelmann, T. (2018).
\newblock {Trace and detect adversarial attacks on CNNs using feature response maps}.
\newblock In {\em Lecture notes in computer science} (Vol. 11081, pp. 346–358). Springer.

\bibitem[\protect\astroncite{Anderson and van Essen}{1987}]{anderson_shifter_1987}
Anderson, C.~H. and van Essen, D.~C. (1987).
\newblock {Shifter circuits: a computational strategy for dynamic aspects of visual processing}.
\newblock {\em Proceedings of the National Academy of Sciences}, 84(17):6297--6301.

\bibitem[\protect\astroncite{Arathorn}{2002}]{Arathorn2002}
Arathorn, D. (2002).
\newblock {\em {Map-seeking circuits in visual cognition: A computational mechanism for biological and machine vision}}.
\newblock Stanford University Press, Stanford, CA.

\bibitem[\protect\astroncite{Carlucci et~al.}{2019}]{carlucci_hallucinating_2019}
Carlucci, F.~M., Russo, P., Tommasi, T., and Caputo, B. (2019).
\newblock {Hallucinating agnostic images to generalize across domains}.
\newblock In {\em Proceedings of the IEEE/CVF International Conference on Computer Vision Workshops} (pp. 3227--3234).

\bibitem[\protect\astroncite{Csurka}{2017}]{csurka_domain_2017}
Csurka, G. (2017).
\newblock {\em {Domain adaptation in computer vision applications}}.
\newblock Springer International Publishing.

\bibitem[\protect\astroncite{Deng et~al.}{2009}]{Deng_2009}
Deng, J., Dong, W., Socher, R., Li, L.-J., Li, K., and Fei-Fei, L. (2009).
\newblock {ImageNet: A large-scale hierarchical image database}.
\newblock In {\em 2009 IEEE Conference on Computer Vision and Pattern Recognition} (pp. 248–255).

\bibitem[\protect\astroncite{Fan et~al.}{2023}]{fan2023towards}
Fan, Q., Segu, M., Tai, Y.-W., Yu, F., Tang, C.-K., Schiele, B., and Dai, D. (2023).
\newblock {Towards robust object detection invariant to real-world domain shifts}.
\newblock In {\em International Conference on Learning Representations (ICLR)}.

\bibitem[\protect\astroncite{Fodor and Pylyshyn}{1988}]{FodorPylyshyn}
Fodor, J.~A. and Pylyshyn, Z.~W. (1988).
\newblock {Connectionism and cognitive architecture: A critical analysis}.
\newblock {\em Cognition}, 28(1):3--71.

\bibitem[\protect\astroncite{Geman et~al.}{1992}]{Geman1992}
Geman, S., Bienenstock, E., and Doursat, R. (1992).
\newblock {Neural networks and the bias/variance dilemma}.
\newblock {\em Neural Computation}, 4(1):1--58.

\bibitem[\protect\astroncite{Greff et~al.}{2020}]{Greff2020}
Greff, K., van Steenkiste, S., and Schmidhuber, J. (2020).
\newblock {On the binding problem in artificial neural networks}.
\newblock {\em arXiv}, 2012.05208.

\bibitem[\protect\astroncite{Hayek}{1952}]{hayek}
Hayek, F.~A. (1952).
\newblock {\em {The sensory order: An inquiry into the foundations of theoretical psychology}}.
\newblock University of Chicago Press, Chicago.

\bibitem[\protect\astroncite{Hebb}{1949}]{hebb_organization_1949}
Hebb, D.~O. (1949).
\newblock {\em {The organization of behavior: A neuropsychological theory}}.
\newblock Wiley, New York.

\bibitem[\protect\astroncite{Hinton and Salakhutdinov}{2006}]{Hinton2006}
Hinton, G.~E. and Salakhutdinov, R. (2006).
\newblock {Reducing the dimensionality of data with neural networks}.
\newblock {\em Science}, 313(5786):504--507.

\bibitem[\protect\astroncite{Hinton and Sejnowski}{1983}]{Hinton1983}
Hinton, G.~E. and Sejnowski, T.~J. (1983).
\newblock {Optimal perceptual inference}.
\newblock In {\em Proceedings of the IEEE Conference on Computer Vision and Pattern Recognition} (pp. 448-453).

\bibitem[\protect\astroncite{Hinton et~al.}{1984}]{hinton_constraint1984}
Hinton, G.~E., Sejnowski, T.~J., and Ackley, D.~H. (1984).
\newblock {Boltzmann machines: Constraint satisfaction networks that learn} (Technical Report CMU-CS-84-119).
\newblock Carnegie Mellon University, Computer Science Department.

\bibitem[\protect\astroncite{Hopfield}{1982}]{Hopfield1982}
Hopfield, J.~J. (1982).
\newblock {Neural networks and physical systems with emergent collective computational abilities}.
\newblock {\em Proceedings of the National Academy of Sciences}, 79(8):2554--2558.

\bibitem[\protect\astroncite{Kingma and Ba}{2015}]{Kingma_Ba_2015}
Kingma, D.~P. and Ba, J. (2015).
\newblock {Adam: A method for stochastic optimization}.
\newblock In {\em International Conference on Learning Representations (ICLR)}.

\bibitem[\protect\astroncite{Krizhevsky and Hinton}{2009}]{krizhevsky2009cifar}
Krizhevsky, A. and Hinton, G.~E. (2009).
\newblock {Learning multiple layers of features from tiny images} (Technical Report).
\newblock University of Toronto.

\bibitem[\protect\astroncite{Leadholm and Stringer}{2022}]{Leadholm2022}
Leadholm, N. and Stringer, S. (2022).
\newblock {Hierarchical binding in convolutional neural networks: Making adversarial attacks geometrically challenging}.
\newblock {\em Neural Networks}, 155:258--286.

\bibitem[\protect\astroncite{LeCun et~al.}{2015}]{lecun_deep_2015}
LeCun, Y., Bengio, Y., and Hinton, G.~E. (2015).
\newblock {Deep learning}.
\newblock {\em Nature}, 521(7553):436--444.

\bibitem[\protect\astroncite{LeCun et~al.}{1989}]{lecun_backpropagation_1989}
LeCun, Y., Boser, B., Denker, J.~S., Henderson, D., Howard, R.~E., Hubbard, W., and Jackel, L.~D. (1989).
\newblock {Backpropagation applied to handwritten zip code recognition}.
\newblock {\em Neural Computation}, 1(4):541--551.

\bibitem[\protect\astroncite{Lecun et~al.}{1998}]{lecun1998gradient}
LeCun, Y., Bottou, L., Bengio, Y., and Haffner, P. (1998).
\newblock {Gradient-based learning applied to document recognition}.
\newblock {\em Proceedings of the IEEE}, 86(11):2278--2324.

\bibitem[\protect\astroncite{Memisevic and Hinton}{2010}]{Memisevic2010}
Memisevic, R. and Hinton, G.~E. (2010).
\newblock {Learning to represent spatial transformations with factored higher-order Boltzmann machines}.
\newblock {\em Neural Computation}, 22(6):1473–1492.

\bibitem[\protect\astroncite{Meyer et~al.}{2025}]{meyer2025hounsfield}
Meyer, B., Sager, P., Abdulkadir, A., Grewe, B.~F., Schuetz, P., Stadelmann, T., and Burn, F. (2025).
\newblock {Hounsfield unit ranges as inductive bias for intra-clinical learning of data-efficient CT segmentation models}.
\newblock In {\em Proceedings of the 12th IEEE Swiss Conference on Data Science (SDS)}. IEEE.

\bibitem[\protect\astroncite{Miconi}{2021}]{miconi_hebbian_2021}
Miconi, T. (2021).
\newblock {Hebbian learning with gradients: Hebbian convolutional neural networks with modern deep learning frameworks}.
\newblock {\em arXiv}, 2107.01729.

\bibitem[\protect\astroncite{Mitchell}{1997}]{learning1997tom}
Mitchell, T. (1997).
\newblock {\em {Machine learning}}.
\newblock McGraw Hill.

\bibitem[\protect\astroncite{Moosavi-Dezfooli et~al.}{2016}]{moosavi-dezfooli_deepfool_2016}
Moosavi-Dezfooli, S.-M., Fawzi, A., and Frossard, P. (2016).
\newblock {DeepFool: A simple and accurate method to fool deep neural networks}.
\newblock In {\em Proceedings of the IEEE Conference on Computer Vision and Pattern Recognition} (pp. 2574--2582).

\bibitem[\protect\astroncite{Olshausen et~al.}{1995}]{Olshausen1995}
Olshausen, B., Anderson, C.~H., and Van Essen, D. (1995).
\newblock {A multiscale dynamic routing circuit for forming size- and position-invariant object representations}.
\newblock {\em Journal of Computational Neuroscience}, 2:45--62.

\bibitem[\protect\astroncite{Olshausen and Field}{2005}]{Olshausen2005}
Olshausen, B.~A. and Field, D.~J. (2005).
\newblock {How close are we to understanding V1?}
\newblock {\em Neural Computation}, 17(8):1665--1699.

\bibitem[\protect\astroncite{Palm}{2013}]{Palm2013}
Palm, G. (2013).
\newblock {Neural associative memories and sparse coding}.
\newblock {\em Neural Networks}, 37:165--171.

\bibitem[\protect\astroncite{Prince}{2023}]{prince2023understanding}
Prince, S. J.~D. (2023).
\newblock {\em {Understanding deep learning}}.
\newblock MIT Press.

\bibitem[\protect\astroncite{Rosenblatt}{1962}]{rosenblatt_principles_1962}
Rosenblatt, F. (1962).
\newblock {\em {Principles of neurodynamics: Perceptrons and the theory of brain mechanisms}}.
\newblock Spartan Books.

\bibitem[\protect\astroncite{Sager et~al.}{2025}]{Sager_2025}
Sager, P.~J., Meyer, B., Yan, P., von Wartburg-Kottler, R., Etaiwi, L., Enayati, A., Nobel, G., Abdulkadir, A., Grewe, B.~F., and Stadelmann, T. (2025).
\newblock {AI agents for computer use: A review of instruction-based computer control, GUI automation, and operator assistants}.
\newblock {\em arXiv}, 2501.16150.

\bibitem[\protect\astroncite{Sager et~al.}{2022}]{sager_unsupervised_2022}
Sager, P., Salzmann, S., Burn, F., and Stadelmann, T. (2022).
\newblock {Unsupervised domain adaptation for vertebrae detection and identification in 3D CT volumes using a domain sanity loss}.
\newblock {\em Journal of Imaging}, 8(8):222.

\bibitem[\protect\astroncite{Simmler et~al.}{2021}]{simmler_survey_2021}
Simmler, N., Sager, P., Andermatt, P., Chavarriaga, R., Schilling, F.-P., Rosenthal, M., and Stadelmann, T. (2021).
\newblock {A survey of un-, weakly-, and semi-supervised learning methods for noisy, missing and partial labels in industrial vision applications}.
\newblock In {\em Proceedings of the 8th Swiss Conference on Data Science (SDS)} (pp. 26--31).

\bibitem[\protect\astroncite{Smolensky}{1987}]{Smolensky1987}
Smolensky, P. (1987).
\newblock Information processing in dynamical systems: Foundations of harmony theory.
\newblock In {\em Parallel distributed processing: Explorations in the microstructure of cognition} (Vol. 1, pp. 194--281).
\newblock MIT Press, Cambridge, MA.

\bibitem[\protect\astroncite{Tsodyks and Feigel'man}{2007}]{Tsodyks2007}
Tsodyks, M. and Feigel'man, M. (2007).
\newblock {The enhanced storage capacity in neural networks with low activity level}.
\newblock {\em EPL (Europhysics Letters)}, 6(1):101.

\bibitem[\protect\astroncite{Tuggener et~al.}{2024}]{tuggener_real_2024}
Tuggener, L., Emberger, R., Ghosh, A., Sager, P., Satyawan, Y.~P., Montoya, J., Goldschagg, S., Seibold, F., Gut, U., Ackermann, P., Schmidhuber, J., and Stadelmann, T. (2024).
\newblock {Real world music object recognition}.
\newblock {\em Transactions of the International Society for Music Information Retrieval}, 7(1):1--14.

\bibitem[\protect\astroncite{Vaswani et~al.}{2017}]{attention2017}
Vaswani, A., Shazeer, N., Parmar, N., Uszkoreit, J., Jones, L., Gomez, A.~N., Kaiser, \L., and Polosukhin, I. (2017).
\newblock {Attention is all you need}.
\newblock In {\em Advances in Neural Information Processing Systems (NeurIPS)} (Vol. 30, pp. 6000–6010).

\bibitem[\protect\astroncite{von~der Malsburg}{1981}]{von_der_malsburg_correlation_1981}
von~der Malsburg, C. (1981).
\newblock {The correlation theory of brain function} (Internal Report 81-2).
\newblock Max-Planck-Institut f\"{u}r Biophysikalische Chemie.

\bibitem[\protect\astroncite{von~der Malsburg}{2018}]{von_der_malsburg_concerning_2018}
von~der Malsburg, C. (2018).
\newblock {Concerning the neuronal code}.
\newblock {\em Journal of Cognitive Science}, 19(4):511--550.

\bibitem[\protect\astroncite{von~der Malsburg}{2023}]{Berlin2023}
von~der Malsburg, C. (2023).
\newblock {Fodor and Pylyshyn's critique of connectionism and the brain as basis of the mind}.
\newblock {\em Frankfurt Institute for Advanced Studies (FIAS)}.

\bibitem[\protect\astroncite{von~der Malsburg}{2024}]{Malsburg2024}
von~der Malsburg, C. (2024).
\newblock {How are segmentation and binding computed and represented in the brain?}
\newblock {\em Cognitive Processing}, 25(1):67--72.

\bibitem[\protect\astroncite{von~der Malsburg et~al.}{2022}]{von_der_malsburg_theory_2022}
von~der Malsburg, C., Stadelmann, T., and Grewe, B.~F. (2022).
\newblock {A theory of natural intelligence}.
\newblock {\em arXiv}, 2205.00002.

\bibitem[\protect\astroncite{Wagemans et~al.}{2012}]{Wagemans2012}
Wagemans, J., Elder, J.~H., Kubovy, M., Palmer, S.~E., Peterson, M.~A., Singh, M., and von~der Heydt, R. (2012).
\newblock {A century of Gestalt psychology in visual perception: I. Perceptual grouping and figure-ground organization}.
\newblock {\em Psychological Bulletin}, 138(6):1172--1217.

\bibitem[\protect\astroncite{Wiskott et~al.}{1997}]{wiskott_face_1997}
Wiskott, L., Fellous, J.-M., Kr\"uger, N., and von~der Malsburg, C. (1997).
\newblock {Face recognition by elastic bunch graph matching}.
\newblock {\em IEEE Transactions on Pattern Analysis and Machine Intelligence}, 19(7):775--779.

\bibitem[\protect\astroncite{Wolfrum et~al.}{2008}]{wolfrum_recurrent_2008}
Wolfrum, P., Wolff, C., L\"{u}cke, J., and von~der Malsburg, C. (2008).
\newblock {A recurrent dynamic model for correspondence-based face recognition}.
\newblock {\em Journal of Vision}, 8(7):34.

\bibitem[\protect\astroncite{Wolpert}{1996}]{NoFreeLunch}
Wolpert, D.~H. (1996).
\newblock {The lack of a priori distinctions between learning algorithms}.
\newblock {\em Neural Computation}, 8(7):1341--1390.

\bibitem[\protect\astroncite{Zeiler and Fergus}{2013}]{Zeiler2013}
Zeiler, M.~D. and Fergus, R. (2013).
\newblock {Visualizing and understanding convolutional networks}.
\newblock {\em arXiv}, 1311.2901.

\bibitem[\protect\astroncite{Zhu and von~der Malsburg}{2004}]{ZHU2004}
Zhu, J. and von~der Malsburg, C. (2004).
\newblock {Maplets for correspondence-based object recognition}.
\newblock {\em Neural Networks}, 17(8):1311--1326.

\end{thebibliography}
\end{document}